\documentclass[sigconf]{aamas}

\usepackage{balance} 
\usepackage{enumitem}
\usepackage{bm}

\usepackage{amssymb}
\usepackage{amsmath}
\usepackage{multirow}  
\usepackage{graphicx}
\usepackage{makecell}
\usepackage{subcaption}
\usepackage{cleveref}
\usepackage{appendix}
\usepackage{algorithm}
\usepackage{algpseudocode}
\usepackage{graphicx}
\usepackage{microtype}
\usepackage{tabularx}
\usepackage{amsthm}
\newtheorem{definition}{Definition}

\makeatletter
\gdef\@copyrightpermission{
  \begin{minipage}{0.2\columnwidth}
   \href{https://creativecommons.org/licenses/by/4.0/}{\includegraphics[width=0.90\textwidth]{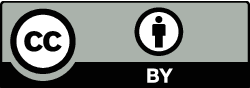}}
  \end{minipage}\hfill
  \begin{minipage}{0.8\columnwidth}
   \href{https://creativecommons.org/licenses/by/4.0/}{This work is licensed under a Creative Commons Attribution International 4.0 License.}
  \end{minipage}
  \vspace{5pt}
}
\makeatother

\setcopyright{ifaamas}
\acmConference[AAMAS '25]{Proc.\@ of the 24th International Conference
on Autonomous Agents and Multiagent Systems (AAMAS 2025)}{May 19 -- 23, 2025}
{Detroit, Michigan, USA}{Y.~Vorobeychik, S.~Das, A.~Nowé  (eds.)}
\copyrightyear{2025}
\acmYear{2025}
\acmDOI{}
\acmPrice{}
\acmISBN{}

\title[Optimizing Action Generation Order in Multi-Agent Reinforcement Learning]{PMAT: Optimizing Action Generation Order in Multi-Agent Reinforcement Learning}

\author{Kun Hu}
\orcid{0009-0002-7686-3606}
\affiliation{
  \institution{National University of Defense Technology}
  \city{Changsha}
  \country{China}
  }
\email{khu@nudt.edu.cn}

\author{Muning Wen}
\orcid{0009-0000-7868-1262}
\affiliation{
  \institution{Shanghai Jiao Tong University}
  \city{Shanghai}
  \country{China}
  }
\email{muningwen@sjtu.edu.cn}

\author{Xihuai Wang}
\orcid{0009-0008-6180-405X}
\affiliation{
  \institution{Shanghai Jiao Tong University}
  \city{Shanghai}
  \country{China}
  }
\email{leoxhwang@sjtu.edu.cn}

\author{Shao Zhang}
\orcid{0000-0002-0111-0776}
\affiliation{
  \institution{Shanghai Jiao Tong University}
  \city{Shanghai}
  \country{China}
  }
\email{shaozhang@sjtu.edu.cn}

\author{Yiwei Shi}
\orcid{0000-0001-6522-9875}
\affiliation{
  \institution{University of Bristol}
  \city{Bristol}
  \country{United Kingdom}
  }
\email{yiwei.shi@bristol.ac.uk}

\author{Minne Li}
\orcid{0009-0004-4768-9164}
\affiliation{
  \institution{Intelligent Game and Decision Lab}
  \city{Beijing}
  \country{China}
  }
\email{lmn.2011@tsinghua.org.cn}

\author{Minglong Li}
\orcid{0000-0001-8239-4573}
\authornote{Correspondence to Minglong Li <liminglong10@nudt.edu.cn> and Ying Wen <ying.wen@sjtu.edu.cn>. Code is available at \url{https://github.com/NUDT-BI-MARL/PMAT}.}
\affiliation{
  \institution{National University of Defense Technology}
  \city{Changsha}
  \country{China}
  }
\email{liminglong10@nudt.edu.cn}

\author{Ying Wen}
\orcid{0000-0003-1247-2382}
\authornotemark[1]
\affiliation{
  \institution{Shanghai Jiao Tong University}
  \city{Shanghai}
  \country{China}
  }
\email{ying.wen@sjtu.edu.cn}

\begin{abstract}
Multi-agent reinforcement learning (MARL) faces challenges in coordinating agents due to complex interdependencies within multi-agent systems. Most MARL algorithms use the simultaneous decision-making paradigm but ignore the action-level dependencies among agents, which reduces coordination efficiency. In contrast, the sequential decision-making paradigm provides finer-grained supervision for agent decision order, presenting the potential for handling dependencies via better decision order management. However, determining the optimal decision order remains a challenge. In this paper, we introduce \textbf{Action Generation with Plackett-Luce Sampling (AGPS)}, a novel mechanism for agent decision order optimization. We model the order determination task as a Plackett-Luce sampling process to address issues such as ranking instability and vanishing gradient during the network training process. AGPS realizes credit-based decision order determination by establishing a bridge between the significance of agents' local observations and their decision credits, thus facilitating order optimization and dependency management. Integrating AGPS with the Multi-Agent Transformer, we propose the \textbf{Prioritized Multi-Agent Transformer (PMAT)}, a sequential decision-making MARL algorithm with decision order optimization. Experiments on benchmarks including StarCraft II Multi-Agent Challenge, Google Research Football, and Multi-Agent MuJoCo show that PMAT outperforms state-of-the-art algorithms, greatly enhancing coordination efficiency.\looseness=-1
\end{abstract}

\keywords{Multi-agent reinforcement learning; Action generation order}

\newcommand{\BibTeX}{\rm B\kern-.05em{\sc i\kern-.025em b}\kern-.08em\TeX}

\begin{document}


\pagestyle{fancy}
\fancyhead{}


\maketitle 


\section{INTRODUCTION}
\label{sec:intro}

\begin{figure}[t]
  \centering
  \vspace{-2pt}
  \includegraphics[width=0.65\linewidth]{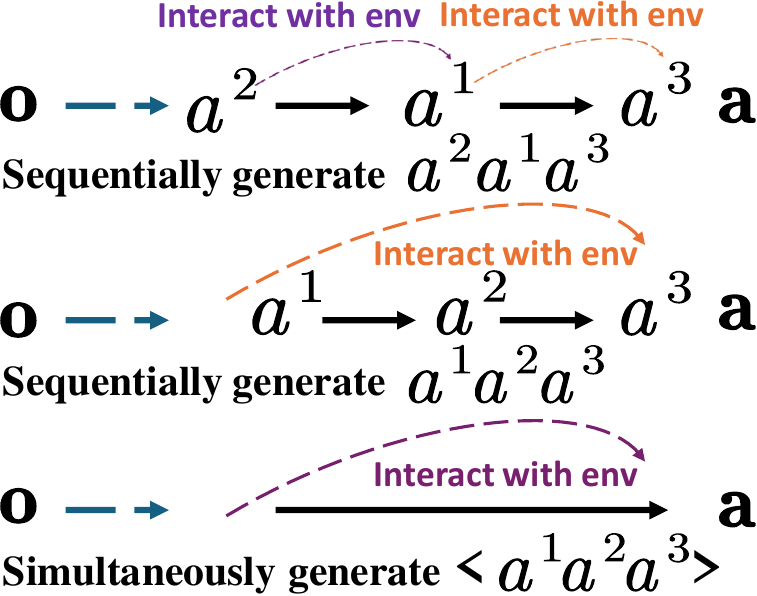}
  \vspace{-4pt}
  \caption{The simultaneous action generation paradigm generates agents' actions concurrently and interacts with the environment once. The sequential action generation paradigm generates agents' actions in an agent-by-agent manner, providing finer-grained supervision for the action generation order. Agents can interact with the environment once per decision or once per iteration under this paradigm.}
  \label{fig:action generation}
  \vspace{-13pt}
\end{figure}

In a multi-agent system (MAS), the optimal action of one agent is often affected by the behavior of others \cite{zhang2021model}, creating complex inter-agent dependency relationships \cite{castelfranchi1992dependence,hannoun1998dependence}. 
Therefore, a key challenge of Multi-Agent Reinforcement Learning (MARL) \cite{zhang2021multi,wang2022model} algorithms is handling the inter-agent dependencies to manage coordination \cite{schumacher2001multi,wang2022model}. 
The dependency relationships among agents necessitate optimizing the decision-making order to achieve optimal team strategies in multi-agent cooperation tasks \cite{castelfranchi1992dependence}.
As illustrated in \Cref{fig:action generation}, MARL algorithms typically employ two decision-making paradigms that generate agents' actions 
either simultaneously \cite{yu2022surprising,wang2023order} or sequentially \cite{wen2022multi}. 
Although simultaneously generating the actions of all agents can facilitate collective learning, it overlooks the potential action-level order dependencies within an MAS.
Consequently, the newly generated action of the concurrent agent may offset the overall performance improvement established by previous agents, resulting in degradation of coordination efficiency \cite{bertsekas2021multiagent}.\looseness=-1

Fortunately, the recent incorporation of sequence models (SMs) in reinforcement learning effectively facilitates sequential decision-making \cite{chen2021decision,janner2021offline}.
In MARL, the utilization of the auto-regressive token generalization mechanism of SMs allows each agent to leverage prior agents' actions during its decision-making process \cite{wen2023large,meng2023offline}, which enables handling dependencies of agents in a sequential manner. 
While the sequential paradigm holds the potential for effective dependency management, identifying the optimal decision order presents a formidable challenge. 
Typically, a fixed or randomized action generation order \cite{wen2022multi} is adopted in this paradigm. 
The fixed-order decision-making scheme is limited by its inability to adapt to agents' dynamically evolving action-level dependencies, which leads to sub-optimal algorithm performance. 
For instance, in a football match, the possession of the ball frequently shifts between players as the game progresses. 
If decisions are made following a fixed order within the team, players might not respond effectively to real-time changes, resulting in poor coordination.\looseness=-1

Although the randomized-order decision-making scheme enables dynamic ordering, learning algorithms may converge to local optima due to random sampling that inadequately represents the solution space or fails to converge to optimal solutions under suboptimal settings \cite{shalev2014understanding}.
Consequently, sequential decision-making in the MARL domain remains fundamentally constrained and less effective without a robust mechanism to determine the optimal decision order.
Hence, there is a pressing need to develop an adaptive decision-ordering mechanism that can effectively manage the dynamic action-level order dependencies among agents.\looseness=-1

While deep learning-based methods have shown remarkable progress in addressing ordering issues \cite{cohen1997learning,pang2017deeprank}, learning the optimal agent decision order still faces several key challenges. On the one hand, directly ordering agents according to their preference scores learned by deep neural networks presents a vanishing gradient issue since the variations of network outputs may not alter the ordering results. On the other hand, when agents declare similar ranking scores, even minor fluctuations in scoring can significantly change the final orderings, thus introducing instability issues.\looseness=-1

To tackle these challenges, we introduce \textbf{Action Generation with Plackett-Luce Sampling (AGPS)}, a Plackett-Luce (P-L) model-based sequential decision-making scheme in MARL. 
Specifically, we formulate the order determination task as an agent-by-agent sampling process and utilize P-L sampling \cite{luce1959individual,plackett1975analysis} in decision order optimization, which facilitates robust and adaptive decision-ordering in multi-agent cooperation tasks. 
Facilitated by AGPS, we propose \textbf{Prioritized Multi-Agent Transformer (PMAT)}, a sequential decision-making MARL algorithm with decision order optimization. We evaluate the proposed algorithm on popular MARL benchmarks including StarCraft II Multi-Agent Challenge \cite{samvelyan2019starcraft}, Google Research Football \cite{kurach2020google}, and Multi-Agent MuJoCo \cite{de2020deep}, where PMAT consistently demonstrates superior task performance compared with several state-of-the-art MARL algorithms.\looseness=-1


\section{RELATED WORK}
\label{sec:related work}
In this section, we introduce several representative state-of-the-art MARL algorithms, covering both the simultaneous and the sequential decision-making paradigms. 
We also discuss the difference between several types of order optimization in MARL.\looseness=-1

\noindent \textbf{Simultaneous Decision-Making MARL Algorithms.} The vast majority of \textit{Centralized Training Decentralized Execution} (CTDE) \cite{oliehoek2008optimal,foerster2018counterfactual} algorithms in MARL adopt a simultaneous decision-making paradigm. 
Here we introduce two representative ones.
MAPPO \cite{yu2022surprising} is a straightforward policy-based approach that endows the policy network of all agents with a shared set of parameters and utilizes agents' aggregated trajectories to facilitate policy optimization. 
HAPPO \cite{kuba2022trust} is a heterogeneous-agent trust-region method that employs a sequential policy update paradigm. 
During an update in HAPPO, the agents randomly choose an update order and update their own policies over the newly updated policies of previous agents. 
Due to the adoption of the simultaneous decision-making paradigm, both MAPPO and HAPPO suffer from potential action conflicts and lack coordination efficiency guarantee.\looseness=-1

\noindent \textbf{Sequential Decision-Making MARL Algorithm.}
To alleviate potential action conflicts and further enhance multi-agent coordination, \citet{wen2022multi} proposed Multi-agent Transformer (MAT), which presents an auto-regressive sequential decision-making MARL algorithm based on the \textit{Transformer} \cite{vaswani2017attention} architecture. MAT successfully transforms multi-agent joint policy optimization into a sequential decision-making process by generating actions in an agent-by-agent manner, which holds the potential for finer-grained supervision and management of inter-agent dependencies.\looseness=-1

\noindent \textbf{Different Order Optimization in MARL.}
The current MARL algorithms have started to focus on the impact of ``order'' on agent-level or batch-level updates, and several solutions have been proposed.
\citet{wang2023order} proposed an agent-by-agent policy optimization method, A2PO, which adopts a semi-greedy agent selection rule to determine agent update order within a single rollout. 
Furthermore, B2MAPO \cite{zhang2024b2mapo} establishes update batches, further enhancing algorithm efficiency to facilitate joint policy optimization in larger-scale agent clusters. 
These studies on sequential update MARL algorithms offer valuable insights, suggesting that decision order optimization can be achieved incrementally on an item-by-item basis.
Unlike these works, this paper focuses on optimizing agent decision order under the sequential decision-making paradigm, which remains an underexplored area in MARL.\looseness=-1


\section{PRELIMINARIES \& BACKGROUND}
\subsection{Cooperative MARL Problem Formulation}
Cooperative MARL problems can usually be modeled as \textit{Markov games} $<\mathcal{N}, \bm{\mathcal{O}}, \bm{\mathcal{A}}, R, P, \gamma>$ \cite{littman1994markov}, where $\mathcal{N} = \{1, \dots, n\}$ is the set of agents, $\bm{\mathcal{O}} = \prod_{i=1}^{n} \mathcal{O}^{i}$ is the joint observation space, $\bm{\mathcal{A}} = \prod_{i=1}^{n} \mathcal{A}^{i}$ is the joint action space, $R: \bm{\mathcal{O}} \times \bm{\mathcal{A}} \longrightarrow \mathbb{R}$ is the joint reward function, $P: \bm{\mathcal{O}} \times \bm{\mathcal{A}} \times \bm{\mathcal{O}} \longrightarrow [0, 1]$ is the transition probability function and $\gamma \in [0, 1)$ is the discount factor. Within each time step, all agents act simultaneously based on their observations. At time step $t \in \mathbb{N}$, each agent $i$ ($i \in \mathcal{N}$) obtains its individual local observation $\mathrm{o}_{t}^{i} \in \mathcal{O}^{i}$ and takes an action $\mathrm{a}_{t}^{i} \in \mathcal{A}^{i}$ according to its own policy $\pi_{i}$, which represents a component of the joint policy $\bm{\pi}$. 

We consider a fully cooperative setting where all agents share the same reward function. When time step $t$ ends, the whole team receives a joint reward $R(\mathbf{o}_{t}, \mathbf{a}_{t})$ and observes $\mathbf{o}_{t+1}$ whose probability distribution is $P(\cdot |\mathbf{o}_{t},\mathbf{a}_{t})$. Following infinitely long times of this process, the multi-agent team finally gains a cumulative return of $R^{\gamma} \triangleq {\textstyle \sum_{t=0}^{\infty}} \gamma^{t} R(\mathbf{o}_{t},\mathbf{a}_{t})$. The observation value and the observation-action value can then be defined as\looseness=-1
\begin{equation}
    V_{\bm{\pi}}(\bm{o}) \triangleq \mathbb{E}_{\mathbf{o}_{1:\infty} \sim P, \mathbf{a}_{0:\infty} \sim \bm{\pi}} [R^{\gamma}|\mathbf{o}_{0}= \bm{o}]
\end{equation}
and
\begin{equation}
    Q_{\bm{\pi}}(\bm{o}, \bm{a}) \triangleq \mathbb{E}_{\mathbf{o}_{1:\infty} \sim P, \mathbf{a}_{1:\infty} \sim \bm{\pi}} [R^{\gamma}|\mathbf{o}_{0}= \bm{o}, \mathbf{a}_{0}= \bm{a}]
\end{equation}
respectively. And the advantage value is defined as
\begin{equation}
    A_{\bm{\pi}}(\bm{o}, \bm{a}) \triangleq Q_{\bm{\pi}}(\bm{o}, \bm{a}) - V_{\bm{\pi}}(\bm{o}).
\end{equation}

\subsection{Multi-Agent Advantage Decomposition}
\label{sec:hypothesis}
In this work, we pay close attention to the impact of action generation order on multi-agent joint advantage improvement in MARL. Before proceeding to our methods, in this section we first introduce existing definitions and theorems as follows:\looseness=-1

\begin{definition}[Multi-Agent Advantage Function \cite{kuba2022trust}]\label{def:MAAF}
     \textit{Let $i_{1:m}$ denote an ordered subset $\{i_1,\ldots,i_m\}$ of $\mathcal{N}$ and $-i_{1:m}$ denote its complement. The multi-agent observation-action value function is defined as
\[
Q_{\bm{\pi}}^{i_{1:m}}(\bm{o},\bm{a}^{i_{1:m}}) \triangleq \mathbb{E}_{\bm{a}^{-i_{1:m}}\sim\bm{\pi}^{-i_{1:m}}}\left[Q_{\bm{\pi}}(\bm{o},\bm{a}^{i_{1:m}},\bm{a}^{-i_{1:m}})\right].
\]
Let $j_{1:k}$ denote another ordered subset of $\mathcal{N}$, such that $i_{1:m} \cap j_{1:k} = \emptyset$. Then, the multi-agent advantage function is defined as
\[
A_{\bm{\pi}}^{i_{1:m}}(\bm{o},\bm{a}^{j_{1:k}},\bm{a}^{i_{1:m}}) \triangleq Q_{\bm{\pi}}^{j_{1:k},i_{1:m}}(\bm{o},\bm{a}^{j_{1:k}},\bm{a}^{i_{1:m}}) - Q_{\bm{\pi}}^{j_{1:k}}(\bm{o},\bm{a}^{j_{1:k}}).
\]}
\end{definition}

\noindent \Cref{def:MAAF} describes the contribution of agents $i_{1:m}$ taking actions $\bm{a}^{i_{1:m}}$ once agents $j_{1:k}$ have taken actions $\bm{a}^{j_{1:k}}$, thus facilitating multi-agent joint policy optimization via the following theorem of\looseness=-1 
\begin{theorem}[Multi-Agent Advantage Decomposition \cite{wen2022multi}]\label{theorem:MAAD}
    Let $i_{1:m}$ be a permutation of agents and $i_k$ denote the $k^{th}$ agent within $i_{1:m}$. Then, for joint observation $\bm{o} = \bm{o} \in \bm{\mathcal{O}}$ and joint action $\bm{a} = \bm{a}^{i_{1:m}} \in \bm{\mathcal{A}}$, the following equation always holds,
\[
A_{\bm{\pi}}^{i_{1:m}}(\bm{o}, \bm{a}^{i_{1:m}}) = \sum_{k=1}^m A_{\bm{\pi}}^{i_k}(\bm{o}, \bm{a}^{i_{1:k-1}}, a^{i_k}).
\]
\end{theorem}

\noindent \Cref{theorem:MAAD} provides an intuitive guide for joint policy optimization within a multi-agent team. It suggests that a sequential optimization of each agent's action contingent upon the actions of preceding agents can finally improve the joint advantage. Hence, one major strength of MARL algorithms adopting the sequential action generation paradigm lies in the potential to ensure that each agent $i_j$ achieves a positive advantage upon the actions $\bm{a}^{i_{1:j-1}}$ of its predecessors via sequential decision-making. As an example, MAT \cite{wen2022multi} utilizes an auto-regressive token generation mechanism to ensure that each agent achieves a positive advantage based on previous agents' actions during the decision-making process.\looseness=-1

\subsection{Multi-Agent Transformer}
Multi-Agent Transformer (MAT) \cite{wen2022multi} is a successful implementation of the \textbf{encoder-decoder} architecture of the \textit{Transformer} \cite{vaswani2017attention} in MARL. The attention mechanism of MAT first encodes agents' observations and actions with a weight matrix calculated by multiplying the embedded queries and keys. Subsequently, representations are calculated by multiplying the weight matrix with the embedded values. In general, the encoder of MAT takes a sequence of observations $(o^{i_1},\ldots,o^{i_n})$ as input and passes them through several computational blocks to generate the corresponding observation representations $(\hat{o}^{i_1},\dots,\hat{o}^{i_n})$. Each of these computational blocks consists of an unmasked self-attention mechanism and a multi-layer perceptron (MLP) to extract the interrelationship among agents. The encoder is trained to approximate the value functions by minimizing the empirical Bellman error of\looseness=-1
\begin{equation}
    L_{\text{Encoder}}(\phi)=\frac{1}{Tn}\sum_{m=1}^{n}\sum_{t=0}^{T-1}\left[R(\mathbf{o}_t,\mathbf{a}_t)+\gamma V_{\bar{\phi}}(\hat{\mathrm{o}}_{t+1}^{i_m})-V_{\phi}(\hat{\mathrm{o}}_t^{i_m})\right]^2,
\end{equation}
where $\phi$ denotes the network parameter and $\bar{\phi}$ denotes the target network parameter. 
The decoder of MAT receives the observation representations output by the encoder. It sequentially generates and passes the embedded actions of agents $\bm{a}^{i_{0:m-1}}$ ($m=1,\ldots n$) through a sequence of decoding blocks, where $a^{i_{0}}$ is an arbitrary symbol indicating the start of decoding. Every decoding block is equipped with a masked self-attention mechanism utilizing triangular matrices to ensure that for each agent $i_j$ attention is computed between the $i^{th}_r$ and the $i^{th}_j$ action heads ($r<j$) so that the sequential scheme can be maintained. The decoding block finally finishes with an MLP and skipping connections, generating a sequence of multi-agent joint action. Parameterized by $\theta$, the decoder is trained to minimize the following clipping PPO \cite{schulman2017proximal} objective of\looseness=-1
\begin{equation}
    \begin{aligned}&L_{\text{Decoder}}(\theta)=-\frac{1}{Tn}\sum_{m=1}^{n}\sum_{t=0}^{T-1}\min\left(\mathrm{r}_{t}^{i_{m}}(\theta)\hat{A}_{t},\mathrm{clip}(\mathrm{r}_{t}^{i_{m}}(\theta),1\pm\epsilon)\hat{A}_{t}\right),\end{aligned}
\end{equation}
where
\begin{equation}
    \mathrm{r}_t^{i_m}(\theta)=\frac{\pi_\theta^{i_m}(\mathrm{a}_t^{i_m}|\hat{\mathbf{o}}_t^{i_{1:n}},\hat{\mathbf{a}}_t^{i_{1:m-1}})}{\pi_{\theta_{\mathrm{old}}}^{i_m}(\mathrm{a}_t^{i_m}|\hat{\mathbf{o}}_t^{i_{1:n}},\hat{\mathbf{a}}_t^{i_{1:m-1}})},
\end{equation}
and $\hat{A}_{t}$ represents an estimate of the joint advantage function. To estimate the joint value function, \textit{generalized advantage estimation} (GAE) \cite{schulman2015high} can be applied with $\hat{V}_t=\frac{1}{n}\sum_{m=1}^{n}V(\hat{\text{o}}_t^{i_m})$.\looseness=-1

\section{DECISION ORDER MATTERS}
\label{sec:joint_adv_opt}

Although the positive-advantage decision-making scheme confers a monotonic improvement guarantee upon MAT, it fails to maximize the joint advantage achieved in each iteration. This stems from lacking effective management of the action-level dependencies among agents. 
Specifically, if the optimal action of agent $i_j$ depends upon the action of agent $i_k$ who plays a pivotal role, enabling $i_k$ to decide prior to $i_j$ provides essential decision-making information for $i_j$, thus holding the potential to enhance the overall team performance (which can also be evidenced by \textit{Example 3}, \cite{bertsekas2021multiagent}).\looseness=-1 

\begin{figure}[tbp]
\vspace{-10pt}
\centering
\begin{subfigure}[b]{0.49\linewidth}
    \includegraphics[width=\linewidth]{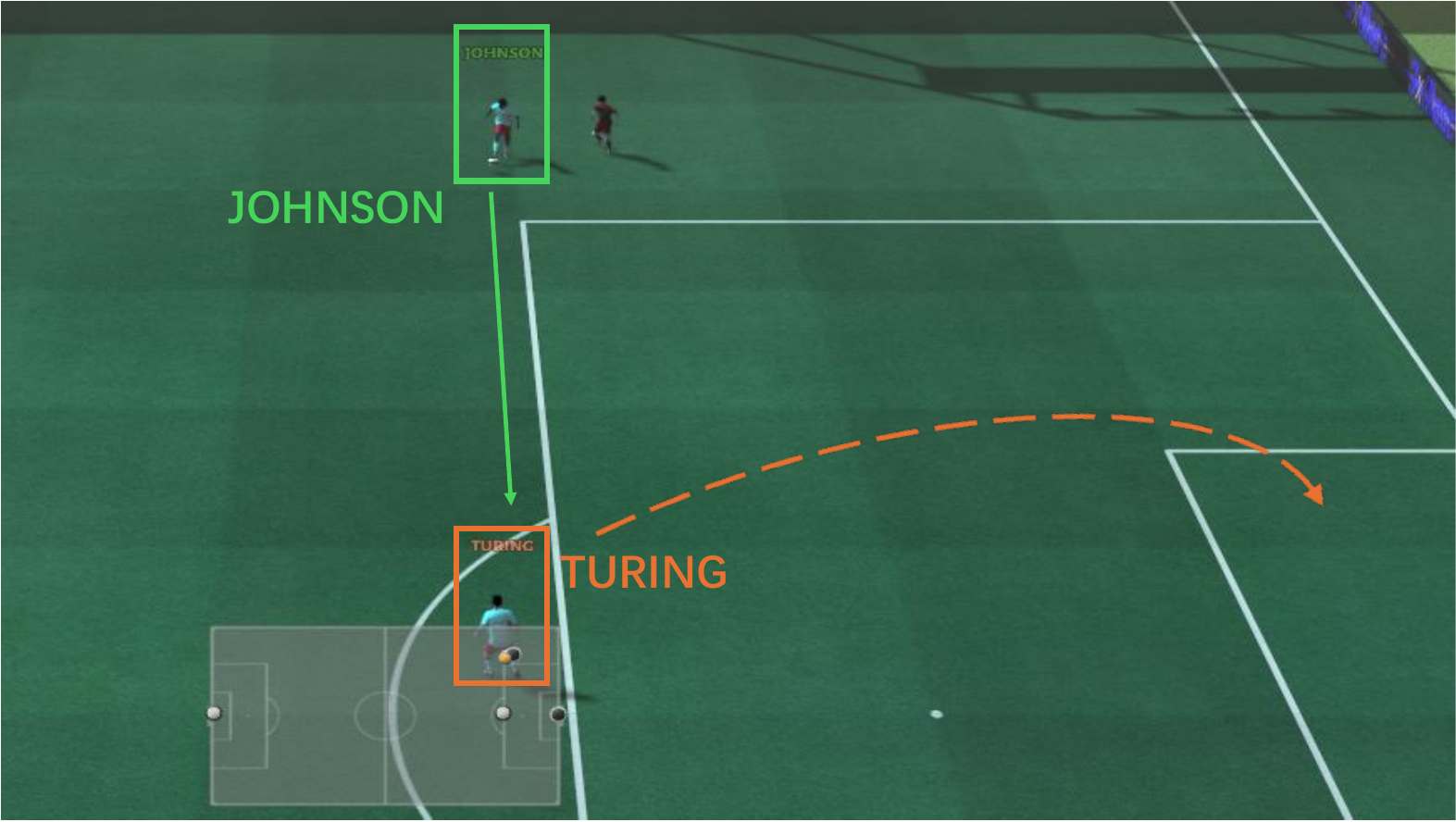}
\end{subfigure}
\hfill 
\begin{subfigure}[b]{0.49\linewidth}
    \includegraphics[width=\linewidth]{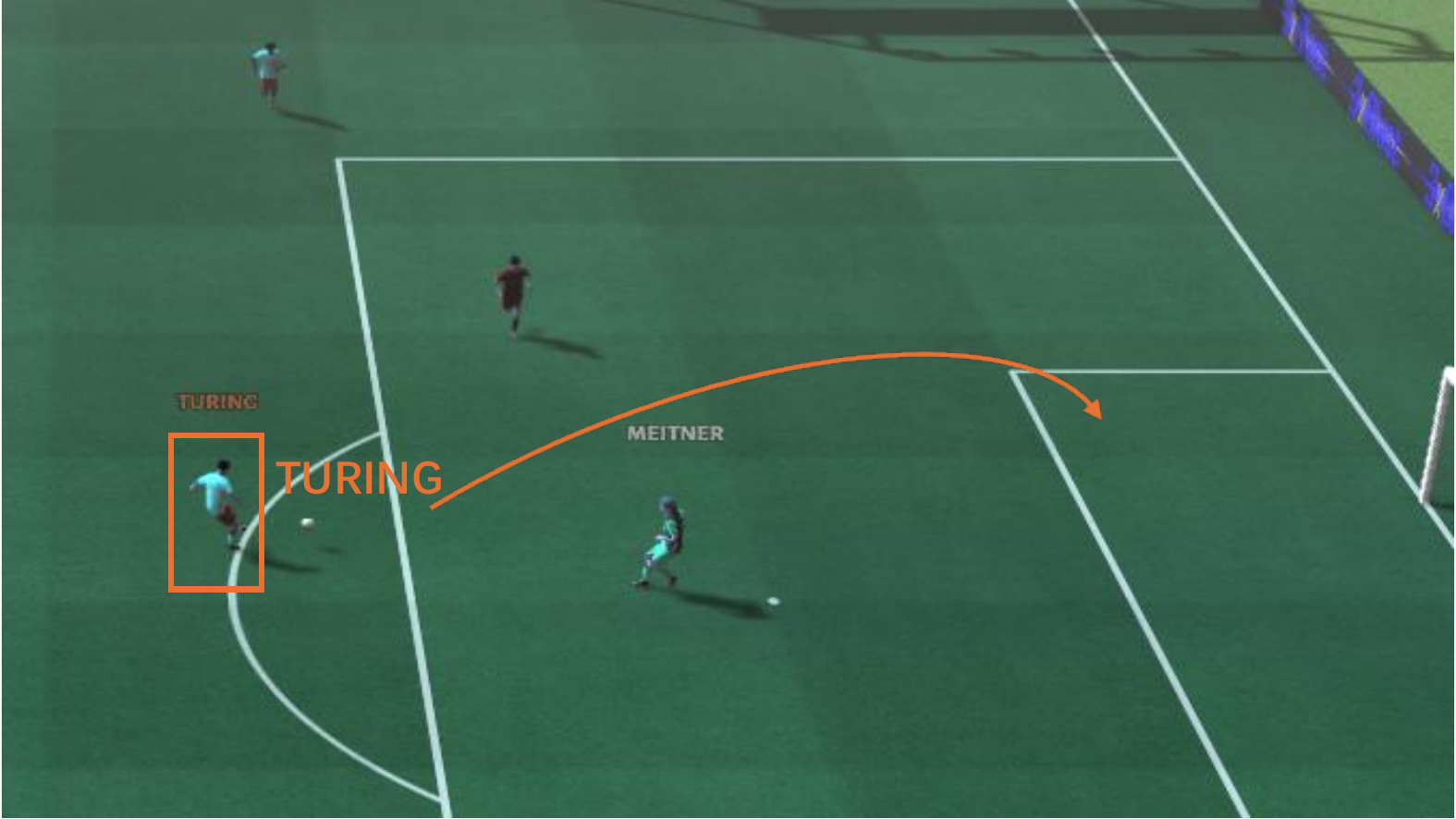}
\end{subfigure}
\vspace{-10pt}
\caption{A multi-agent cooperation scenario taken from Google Research Football. Player JOHNSON passes the ball to his partner TURING who has a favorable shooting angle (left), and TURING converts the shot into a goal (right).}
\label{fig:pass and shoot}
\vspace{-12pt}
\end{figure}

As an illustrative example, in \Cref{fig:pass and shoot}, we take two frames from the \textit{academy pass and shoot with keeper} scenario of Google Research Football \cite{kurach2020google}, where the local observation of Player TURING exhibits prior significance due to his advantageous positioning for scoring a goal. In this case, TURING is allowed to decide first and he decides to take a shot as illustrated in \Cref{fig:pass and shoot}, where decisions are specially plotted in dashed arrows. Subsequently, TURING's teammate JOHNSON decides to pass the ball, considering both TURING's decision (shoot) and his own observation (the position of TURING and the opponent player etc.). In such a prioritized sequential decision-making manner, subsequent players hold the potential to recognize the intentions of preceding players who possess more significant local observations and align their actions with these predecessors in an efficient way, thus facilitating the emergence of collaborative behavior among agents and enhancing the overall task performance of the whole team. Later, we will build upon this insight to introduce a sequential action generation scheme that optimizes the agent decision order according to the significance of their local observations to the joint advantage.\looseness=-1

To further discuss the impact of decision order, we define the \textit{action generation order} $\sigma$ as a permutation of $\mathcal{N}$, which induces the multi-agent joint policy (contingent upon this order) as\looseness=-1
\begin{equation}
    \bm{\pi}^{\sigma}_{i_{1:n}} = \pi_{i_1}(a^{i_1}|\bm{o}, \sigma) \cdot \pi_{i_2}(a^{i_2}|\bm{o}, \sigma, a^{i_1}) \cdot \ldots \cdot \pi_{i_n}(a^{i_n}|\bm{o}, \sigma, a^{i_1}, \ldots, a^{i_{n-1}}).
\end{equation}
Then, we define the \textit{optimal action generation order} as
\begin{definition}[Optimal Action Generation Order]\label{def:OAGO}
    Given a group of agents marked as $\{1,\dots,n\}$, if an action generation order $\sigma^{*} = \{i_1, i_2, \dots, i_n\}$ satisfies $A_{\bm{\pi}^{\sigma^*}}^{1:n}(\bm{o}, \bm{a}^{1:n}_{\sigma^*}) \ge A_{\bm{\pi}^{\sigma}}^{1:n}(\bm{o}, \bm{a}^{1:n}_{\sigma})$ for any $\sigma \neq \sigma^*$, then we define $\sigma^{*}$ as the Optimal Action Generation Order.
\end{definition}

Applying different action generation orders can result in distinct joint action, ultimately affecting the multi-agent joint advantage achieved in each iteration. Hence, optimizing agent decision order is significant to joint advantage optimization in MARL. \Cref{def:OAGO} defines the optimal agent decision order, utilizing the joint advantage value as feedback signal. Our goal is to identify and utilize an optimized decision-making order as the action generation order in each round of the multi-agent sequential decision-making process. By utilizing this refined order, we aim to optimize the joint advantage achieved in each iteration of multi-agent joint action, thereby enhancing the efficiency and task performance of MARL algorithms.\looseness=-1


\section{DECISION ORDER OPTIMIZATION}

In this section, we first discuss the challenges of learning the optimal decision order and highlight the advantages of utilizing the Plackett-Luce (P-L) model for decision order optimization. We then introduce AGPS, a P-L model-based sequential decision-making mechanism. Additionally, we also present a practical MARL algorithm that serves as an application instantiation of AGPS.\looseness=-1

\subsection{Decision Ordering as a Ranking Task}
A straightforward approach to optimizing the agent decision order involves evaluating all permutations of \( n \) agents within each iteration of the sequential action generation process, with the objective of identifying the specific permutation that can maximize the joint advantage value. 
While this method exhibits favorable interpretability and can guarantee optimal orderings, the primary limitation stems from its factorial search complexity ($\mathcal{O}(n!)$), which significantly increases the computational cost and limits its applicability in large-scale multi-agent systems.
Inspired by the football case depicted in \Cref{sec:joint_adv_opt},  we propose to leverage the potential correlation between the optimal decision order and the preference scoring of agents' local observations to address this challenge. Specifically, we model the decision-ordering task as a label ranking problem, enabling the application of parametric probabilistic models \cite{cheng2010label} and deep learning-based optimization techniques.\looseness=-1

Learning to rank is a fundamental problem in the domain of machine learning \cite{cao2007learning, xia2008listwise}, where deep learning-based methods have witnessed widespread applications \cite{severyn2015learning}. By establishing a scoring network to evaluate the preference scores of agents' local observations and ranking them accordingly, an optimized decision sequence can be derived in a computationally efficient manner. However, this approach also exhibits several limitations when implemented in multi-agent systems. Firstly, the output variations of the scoring network do not necessarily convert into adjustments in the final rankings, which can potentially induce the vanishing gradient issue during the neural network training process. Secondly, generating agent decision order in a deterministic manner can introduce instability in that, when the individual scores are similar, minor discrepancies in scoring can produce substantial alterations in the final rankings. 
To address these challenges, we model the decision-ordering task as a multi-step sampling process and propose a P-L model-based approach that facilitates decision-credit allocation and decision-order optimization in multi-agent sequential decision-making.\looseness=-1

\subsection{Plackett-Luce Sampling}
The Plackett-Luce model derives its name from independent work by Plackett \cite{plackett1975analysis} and Luce \cite{luce1959individual}, which has found extensive applications in various real-world tasks like horse-racing \cite{plackett1975analysis}, document ranking \cite{cao2007learning} and information retrieval \cite{guiver2009bayesian}, etc. A P-L model is parameterized by an $n$-length vector $\bm{v}=(v_1,\dots,v_n)$ where $v_i>0$ represents the preference score associated with each object $i$. The probability of sampling an ordered permutation $\sigma^{(n)}=(\sigma(1),\ldots,\sigma(n))$ from a P-L distribution can be written as\looseness=-1
\begin{equation}
    \label{form:P-L prob}
    \mathrm{P}(\sigma^{(n)} \mid \bm{v})=\prod_{i=1}^n\frac{v_{\sigma(i)}}{v_{\sigma(i)}+v_{\sigma(i+1)}+\ldots+v_{\sigma(n)}}.
\end{equation}

The P-L model extends the Bradley-Terry (BT) model suggested by \citet{bradley1952rank}, which is renowned for its application in the domain of pairwise comparisons, to model item preferences as sampling probabilities. Specifically, the BT model specifies the probability that ``$\sigma(i)$ wins against $\sigma(j)$'' in terms of\looseness=-1
\begin{equation}
    \label{form:asymmetric relation}
    \mathrm{P}(\sigma(i)\succ\sigma(j))=\frac{v_{\sigma(i)}}{v_{\sigma(i)}+v_{\sigma(j)}},
\end{equation}
where $\succ$ denotes the asymmetric relation which indicates that $\sigma(i)$ precedes $\sigma(j)$ in ordering. 
Derived from the BT model, the expected ordering generated from a P-L sampling process satisfies\looseness=-1 
\begin{equation}
    \begin{aligned}\sigma^{(n)}&=[\sigma(1),\sigma(2),\dots,\sigma(n)],\\&\mathrm{s.t.} \quad \forall(\sigma(i),\sigma(j))\text{, }i<j \Rightarrow v_i \geq v_j. \end{aligned}
\end{equation}
P-L sampling provides a probabilistic understanding of preference structures, addressing key challenges such as instability and vanishing gradient associated with the ordering process. Specifically, P-L sampling decomposes the ranking task of \(n\) objects as a sequence of \(n-1\) independent selection stages, wherein each stage involves choosing the next top-scoring item from the remaining alternatives. This sequential mechanism ensures that objects with similar preference scores are assigned comparable selection probabilities, thereby reducing the sensitivity to minor score fluctuations and enhancing the robustness of ranking. Additionally, this method effectively converts the variations in preference scores output by neural networks into adjustments in the final rankings, thus mitigating the vanishing gradient issue in the neural network training process. Furthermore, P-L sampling has been shown to be computationally efficient \cite{oosterhuis2021computationally} and offers various optimization opportunities, indicating potential scalability in large-scale multi-agent systems.\looseness=-1

\subsection{Action Generation with P-L Sampling}
We utilize P-L sampling to optimize the action generation order within an MAS. To handle non-negative constraints, we parameterize the multi-agent P-L distribution using logarithmic parameters \(\bm{z} = (z_1, \ldots, z_n)\). The probability of obtaining the optimal action generation order \(\sigma^*\) can then be derived as\looseness=-1
\begin{equation}
    P(\sigma^{*}|\bm{z})=\prod_{i=1}^{n-1}\frac{\exp z_{\sigma^*(i)}}{\sum_{j=i}^n\exp z_{\sigma^*(j)}}.
\end{equation}
Utilizing the sequence-related joint advantage value \(A_{\bm{\pi}^{\sigma}}^{i_{1:n}}\) (abbreviated as \(A(\sigma)\)) as feedback signal, the parameters \(\bm{z}\) are learned through a scoring network that outputs the preference scores associated with agents' local observations. Specifically, let \(\mathbb{S}_{\mathcal{N}}\) denote the space of \(n\)-agent permutations, the expectation-form objective function of the agent decision order optimization problem can be formulated as
\begin{equation}
\label{form:opti-obj}
J(\bm{z})=\mathbb{E}_{\sigma \in \mathbb{S}_{\mathcal{N}}}\begin{bmatrix}A(\sigma)\end{bmatrix}=\sum_{\sigma \in \mathbb{S}_{\mathcal{N}}}A(\sigma) P(\sigma|\bm{z}).
\end{equation}
For numerical computation, the gradient of \Cref{form:opti-obj} can be estimated via \textit{Monte Carlo approximation} as
\begin{equation}
\label{log-grad-PL}
\nabla_{\bm{z}}J(\bm{z})=\mathbb{E}_{\sigma \in \mathbb{S}_{\mathcal{N}}}\left[A(\sigma)\nabla_{\bm{z}}\log P(\sigma|\bm{z})\right] \approx \frac{1}{N}\sum_{i=1}^{N}A(\sigma_i)\nabla_{\bm{z}}\log P(\sigma_i|\bm{z}).
\end{equation}
The $i^{th}$ partial derivative of the log-likelihood $\log P(\sigma|\bm{z})$ with respect to $z_{\sigma(i)}$ in \Cref{log-grad-PL} can be calculated by
\begin{equation}
    \frac{\partial\log P(\sigma|\bm{z})}{\partial z_{\sigma(i)}}=1-\exp (z_{\sigma(i)})\sum_{k=1}^i\frac{1}{\sum_{j=k}^n\exp (z_{\sigma(j)})},
\end{equation}
and the full gradient $\nabla_{\bm{z}}\log P(\sigma|\bm{z})$ can be generated within $\mathcal{O}(n)$ timesteps \cite{ceberio2023model}, thus demonstrating superior efficiency.\looseness=-1

We designate the proposed sequential action generation mechanism as \textbf{Action Generation with P-L Sampling (AGPS)}. AGPS allows agents whose observations contribute more significantly to the multi-agent joint advantage to be granted higher decision-making priority. In this manner, subsequent agents can effectively perceive the decisions of their predecessors and offer proactive cooperation during their decision-making process, thereby enhancing the overall coordination within a multi-agent team.\looseness=-1
\begin{figure*}[htbp]
	\includegraphics[width=0.90\linewidth]{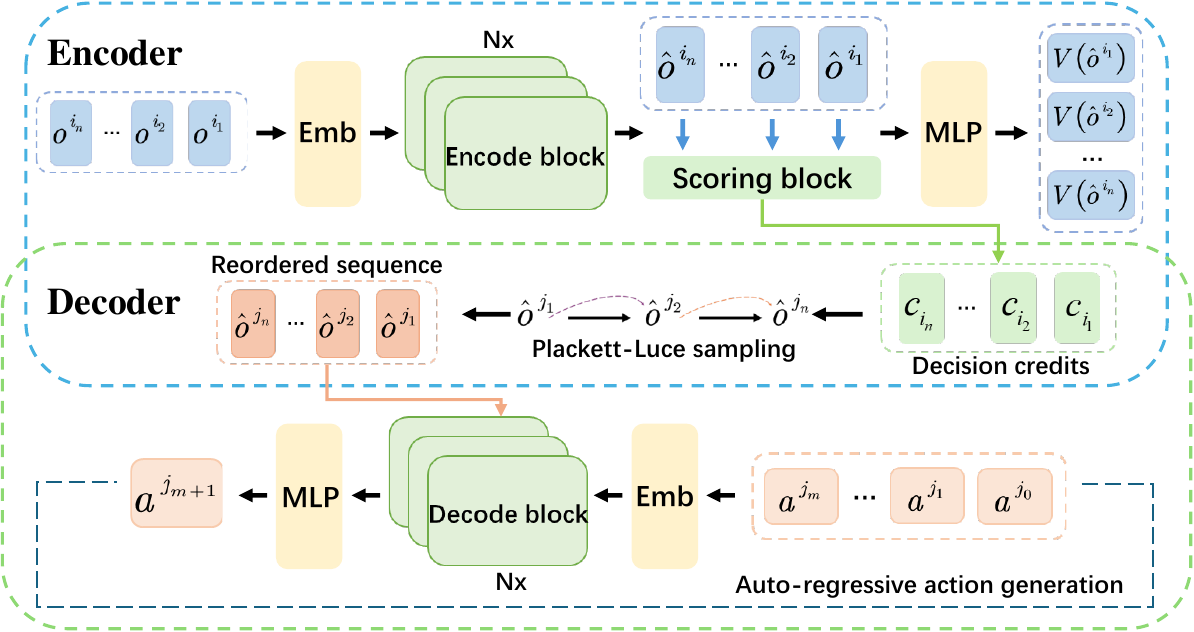}
	\caption{The overall framework of the proposed \textit{Prioritized Multi-Agent Transformer}. The encoder processes agents' local observations at each time step, transforming them into high-level representations. The observation representations are subsequently fed into the scoring block to generate individual preference scores, referred to as decision credits. P-L sampling is then conducted based on the scoring to compute the action generation order. The representations are reordered prior to being fed into the decoder which sequentially generates agents' actions in accordance with this reordered sequence.\looseness=-1}
	\label{fig:pmat}
 \vspace{-8pt}
\end{figure*}
 
\subsection{Practical Algorithm}
As illustrated in \Cref{fig:action generation}, sequential decision-making MARL algorithms can be implemented via single or multiple interactions with the environment. While the latter paradigm benefits from timely feedback, it suffers from degradation in computational efficiency due to numerous interactions within each iteration of multi-agent joint action. Fortunately, recent advancements in auto-regressive SMs have provided fresh insights into the single-interaction paradigm, facilitating efficient sequential decision-making MARL algorithms based on the \textit{Transformer} \cite{vaswani2017attention} architecture.\looseness=-1

\paragraph{\textbf{Prioritized Multi-Agent Transformer}}
We demonstrate that AGPS can be effectively integrated with MAT for performance enhancement. On the one hand, the observation representations output by the encoder of MAT synthesize not only the local observations of individual agents but also the high-level inter-agent relationships \cite{wen2022multi}, thus providing informative signals for preference scoring. On the other hand, the masked self-attention mechanism of MAT's decoder inherently promotes efficient sequential action generation. Hence, via integrating P-L sampling with the encoder-decoder architecture of MAT, an effective bridge between the local observation representations (as outputs of the encoder) and the decision orderings (as inputs of the decoder) can be established, which facilitates auto-regressive sequential action generation and ultimately leads to the proposed \textbf{Prioritized Multi-Agent Transformer (PMAT)} as illustrated in \Cref{fig:pmat}.\looseness=-1 

As shown in \Cref{fig:pmat}, the scoring block consists of an MLP $\Phi$ parameterized by $\varphi$, which takes agents' local observation representations $(\hat{o}^{i_1},\ldots,\hat{o}^{i_n})$ as input and yields the corresponding preference scores $(c_{i_1},\dots,c_{i_n})$. Specifically, the scoring network is trained to minimize the following clipping objective of\looseness=-1
\begin{equation}
    L_{\mathrm{Ranking}}(\varphi)=-\frac1{T}\sum_{t=0}^{T-1}\min\left(\mathrm{r}_t^{\sigma}(\varphi)\hat{A}_t,\mathrm{clip}(\mathrm{r}_t^{\sigma}(\varphi),1\pm\epsilon)\hat{A}_t\right),
\end{equation}
where
\begin{equation}
    \mathrm{r}_t^{\sigma}(\varphi)=\frac{P(\sigma|\Phi_{\varphi}(\hat{\mathbf{o}}_t^{i_{1:n}}))}{P(\sigma|\Phi_{\varphi_{old}}(\hat{\mathbf{o}}_t^{i_{1:n}}))},
\end{equation}
$\hat{A}_t$ estimates the sequence-related joint advantage, and $P(\sigma|\cdot)$ represents the probability of obtaining the current order $\sigma=(j_1,\dots,j_n)$ via P-L sampling. 
Before being passed into the decoder, the original observation representations $(\hat{o}^{i_1},\ldots,\hat{o}^{i_n})$ are reordered according to this order as $(\hat{o}^{j_1},\ldots,\hat{o}^{j_n})$. Meanwhile, $\sigma=(j_1,\dots,j_n)$ also serves as the auto-regressive action generation order within the decoder, thus realizing agent decision order optimization.\looseness=-1 

Notably, the sampling paradigm differs slightly between the training and inference stages. During the training stage, P-L sampling is carried out in a non-deterministic manner since introducing randomness can facilitate generalization capability and mitigate the risk of overfitting specific training instances. During the inference stage, however, deterministic sampling is carried out for performance enhancement. Besides, in the training stage, the output of all actions $\bm{a}^{i_{1:n}}$ can be computed with parallel acceleration in the sense that $\bm{a}^{i_{1:n-1}}$ have already been collected and stored in the replay buffer. In contrast, during the inference stage, each action $a^{i_m}$ has to be inserted back into the decoder auto-regressively to generate the following action $a^{i_{m+1}}$ in a sequential manner.\looseness=-1
\begin{figure*}[h]
  \centering
  \begin{subfigure}[b]{0.33\textwidth}
    \includegraphics[width=\textwidth]{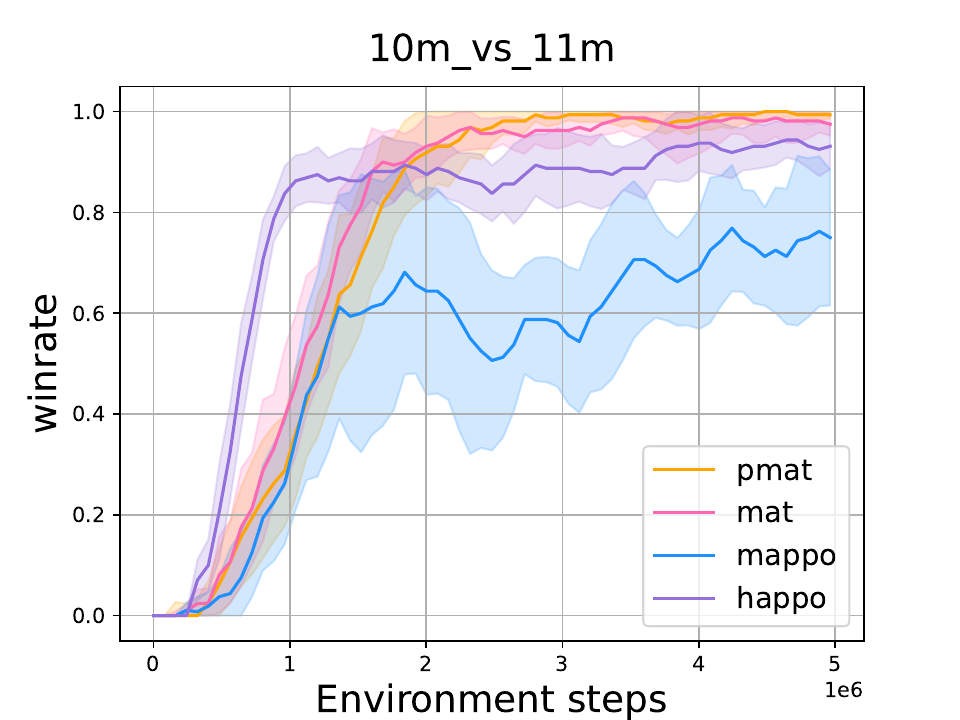}
    \caption{10m vs 11m (SMAC)}
    \label{fig:10m_vs_11m}
  \end{subfigure}
  \hfill
  \begin{subfigure}[b]{0.33\textwidth}
    \includegraphics[width=\textwidth]{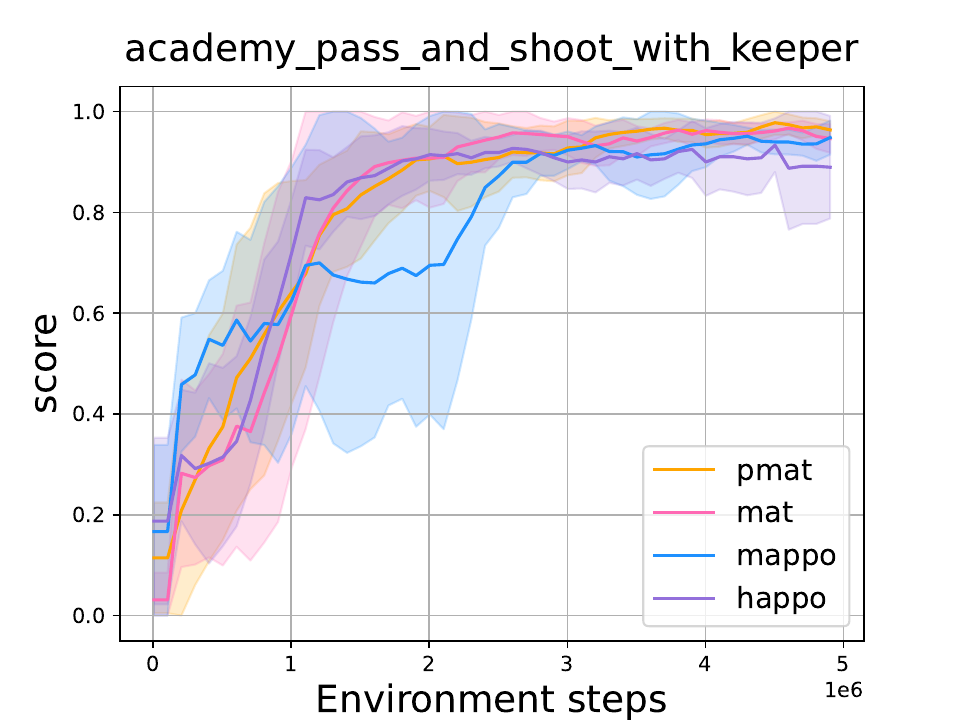}
    \caption{academy pass and shoot with keeper (GRF)}
    \label{fig:academy_pass_and_shoot_with_keeper}
  \end{subfigure}
  \hfill
  \begin{subfigure}[b]{0.33\textwidth}
    \includegraphics[width=\textwidth]{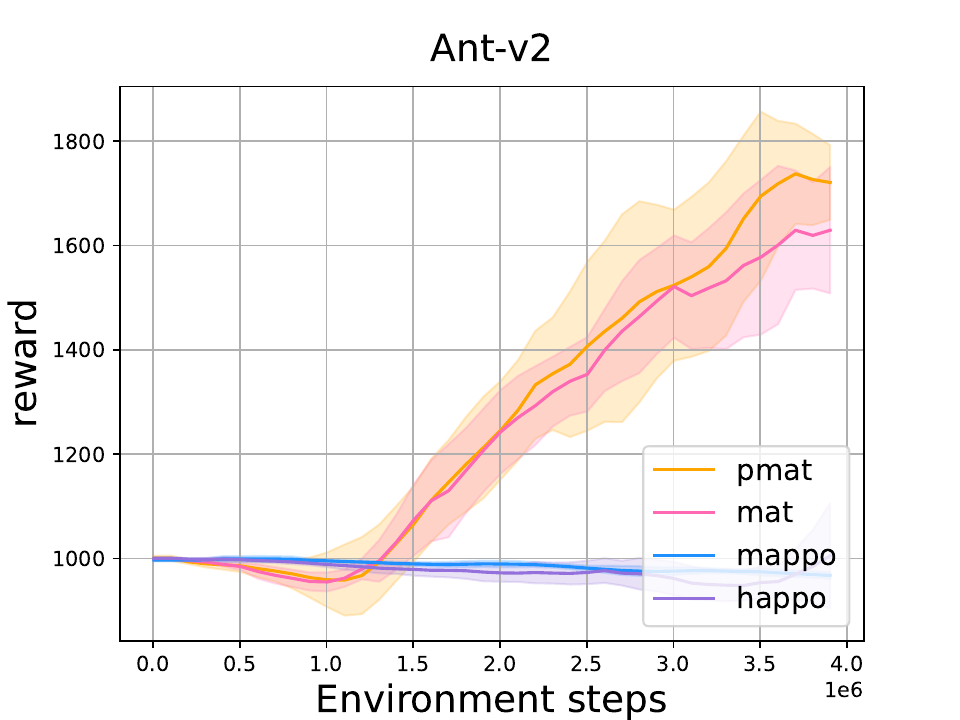}
    \caption{8×1 agent Ant (MA MuJoCo)}
    \label{fig:8×1 agent Ant}
  \end{subfigure}
  \\
  \vspace{1em}
  \begin{subfigure}[b]{0.33\textwidth}
    \includegraphics[width=\textwidth]{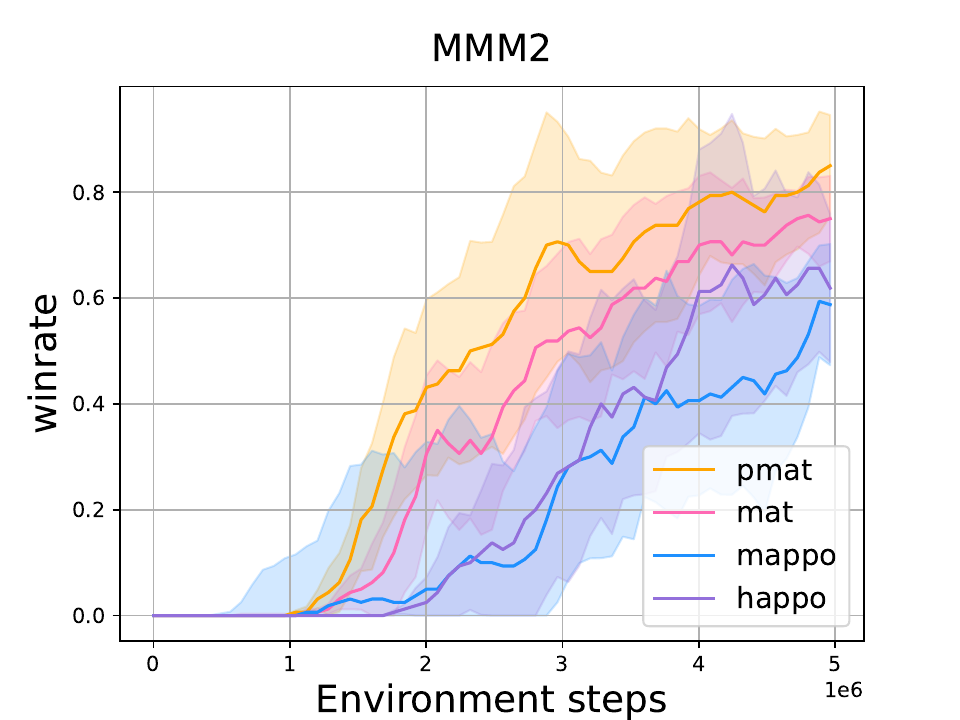}
    \caption{MMM2 (SMAC)}
    \label{fig:MMM2}
  \end{subfigure}
  \hfill
  \begin{subfigure}[b]{0.33\textwidth}
    \includegraphics[width=\textwidth]{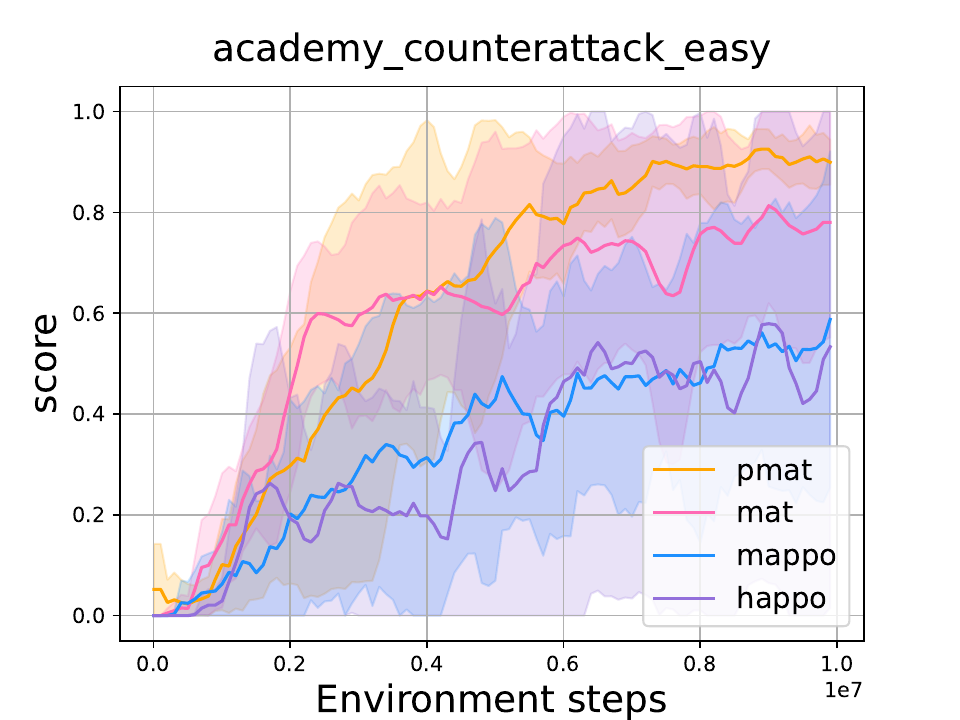}
    \caption{academy counterattack easy (GRF)}
    \label{fig:academy_counterattack_easy}
  \end{subfigure}
  \hfill
  \begin{subfigure}[b]{0.33\textwidth}
    \includegraphics[width=\textwidth]{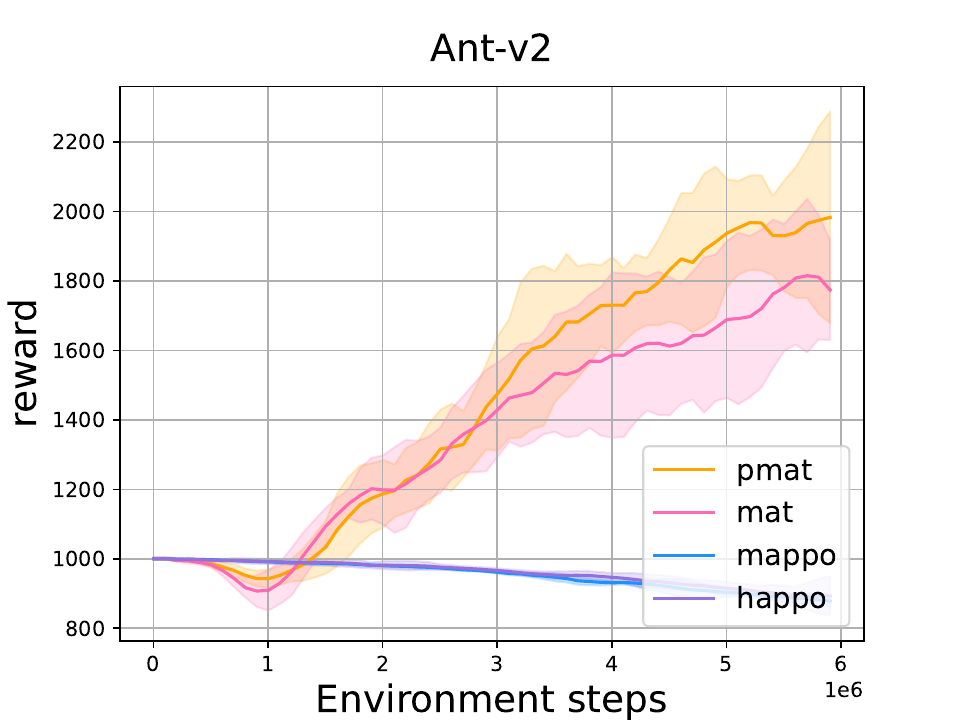}
    \caption{4×2 agent Ant (MA MuJoCo)}
    \label{fig:4×2 agent Ant}
  \end{subfigure}
  \vspace{-8pt}
  \caption{Experimental results in StarCraft II, Google Research Football, and Multi-Agent MuJoCo.}
  \label{fig:experimental results}
\end{figure*}
\begin{figure*}[h]
  \centering
  \begin{subfigure}[b]{0.245\textwidth}
    \includegraphics[width=\textwidth]{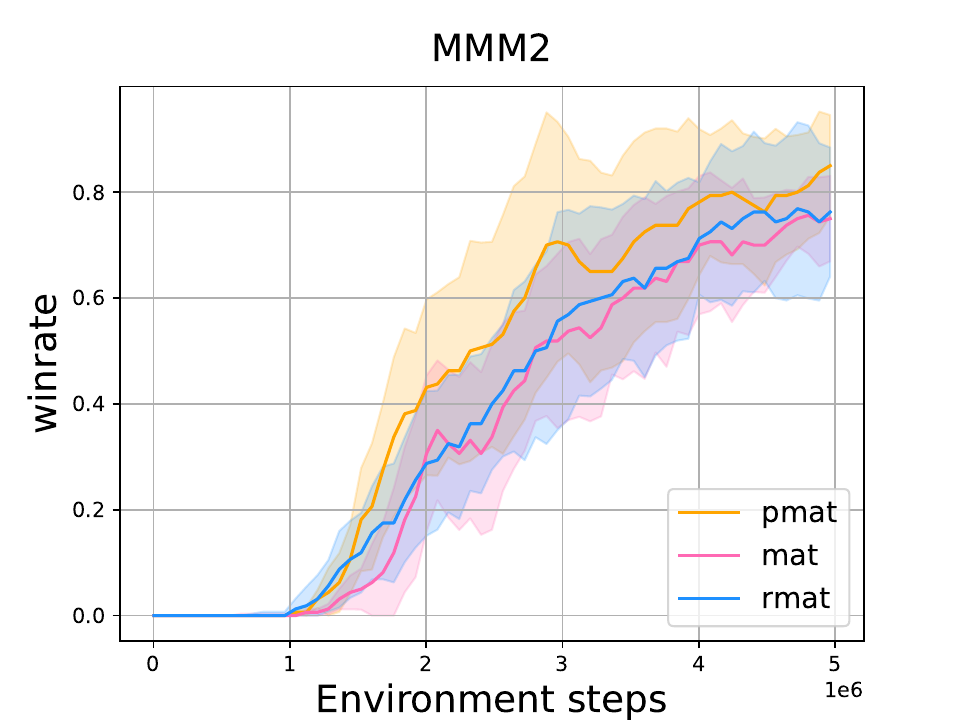}
    \label{fig:abla_MMM2}
  \end{subfigure}
  \hfill
  \begin{subfigure}[b]{0.245\textwidth}
    \includegraphics[width=\textwidth]{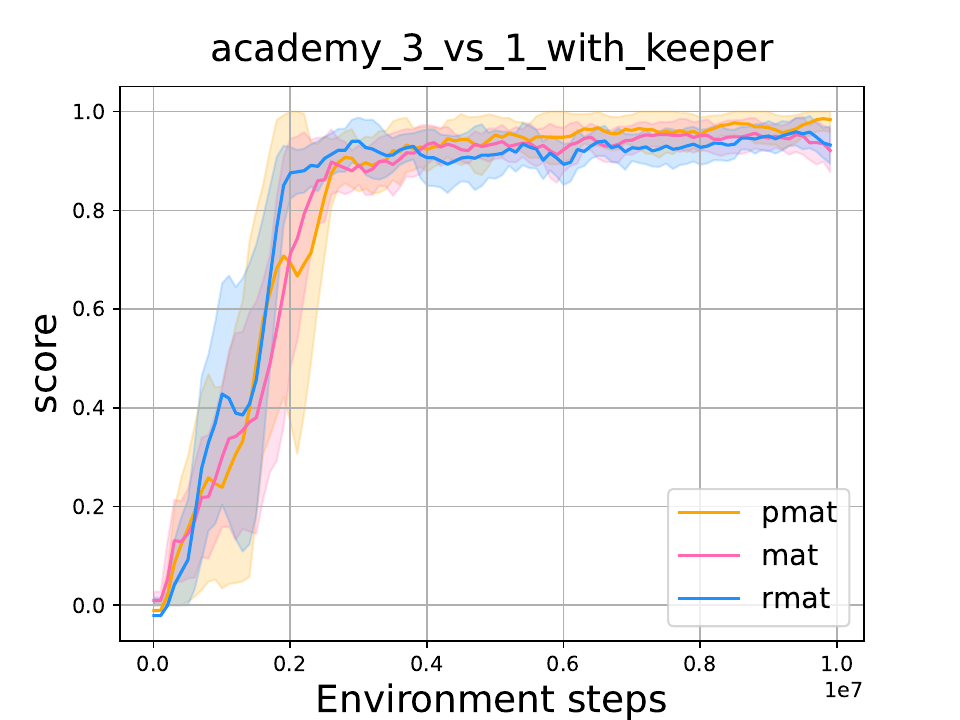}
    \label{fig:abla_academy_3_vs_1_with_keeper}
  \end{subfigure}
  \hfill
  \begin{subfigure}[b]{0.245\textwidth}
    \includegraphics[width=\textwidth]{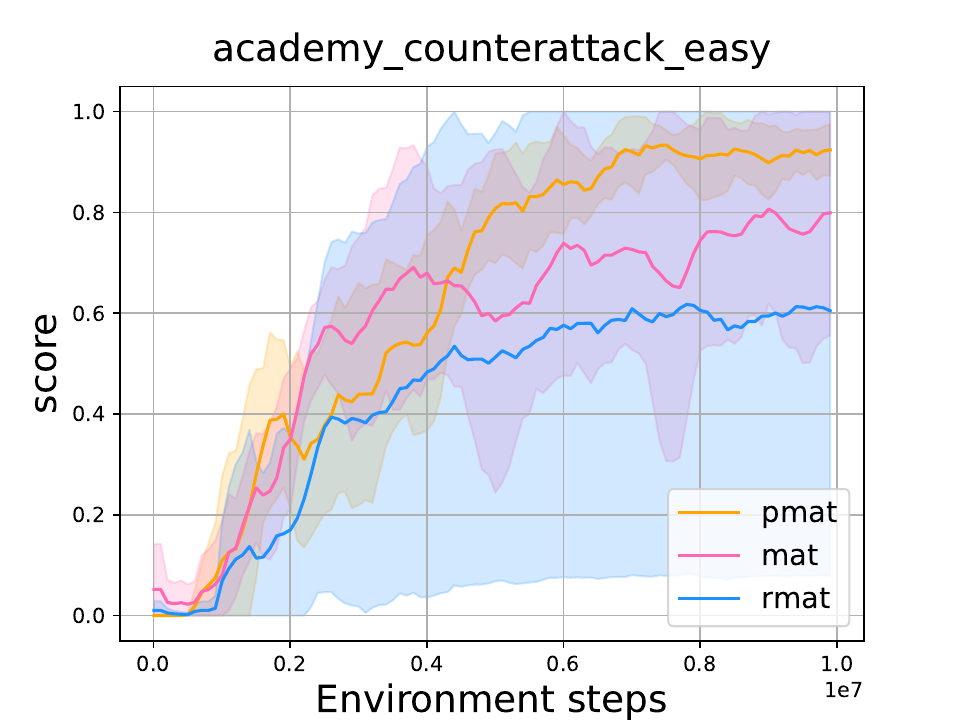}
    \label{fig:abla_academy_counterattack_easy}
  \end{subfigure}
  \hfill
  \begin{subfigure}[b]{0.245\textwidth}
    \includegraphics[width=\textwidth]{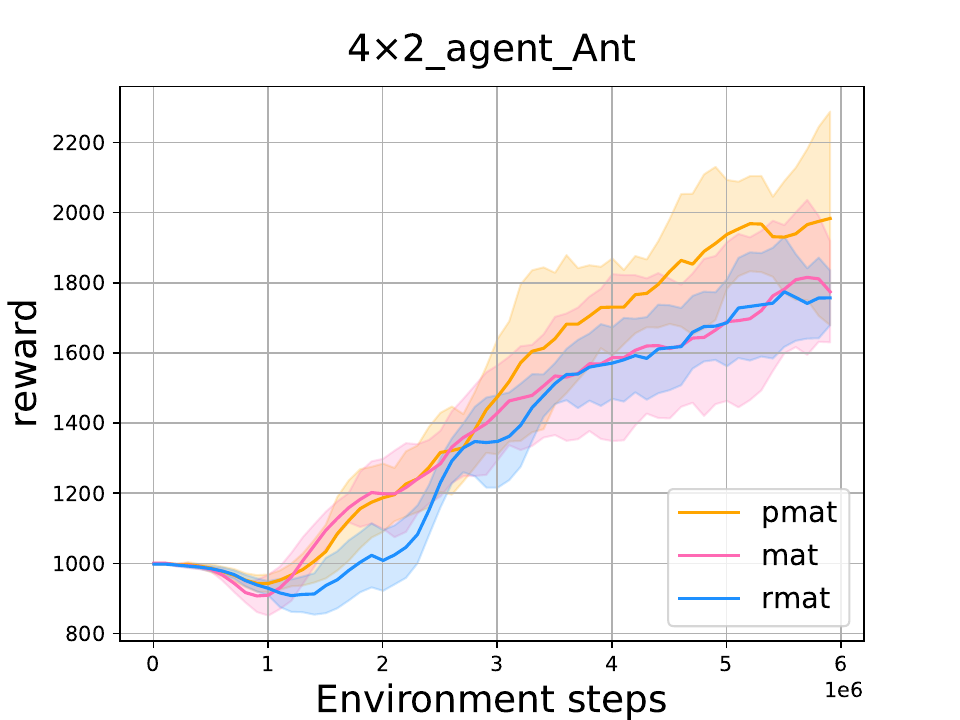}
    \label{fig:abla_4×2_agent_Ant}
  \end{subfigure}
  \vspace{-8pt}
  \caption{Ablation results in StarCraft II, Google Research Football, and Multi-Agent MuJoCo.}
  \label{fig:ablation results}
  \vspace{-8pt}
\end{figure*}


\section{EXPERIMENTS}
In this section, we evaluate the effectiveness of the proposed AGPS and its application instantiation, PMAT, within popular MARL benchmarks.  
We compare PMAT with advanced MARL methods including MAT \cite{wen2022multi}, MAPPO \cite{yu2022surprising}, and HAPPO \cite{kuba2022trust}.\looseness=-1

\subsection{Experimental Environments}
In this work, we evaluate our method within the following three MARL benchmarks: StarCraft II Multi-Agent Challenge (referred to as SMAC) \cite{samvelyan2019starcraft}, Google Research Football (referred to as GRF) \cite{kurach2020google} and Multi-Agent MuJoCo (referred to as MA MuJoCo) \cite{de2020deep}.\looseness=-1 

\noindent \textbf{StarCraft II Multi-Agent Challenge}. SMAC \cite{samvelyan2019starcraft} is an open-source research environment designed to evaluate MARL algorithms based on the StarCraft II game engine. 
SMAC simulates complex scenarios and varying unit types with real-time multi-agent interactions, enabling comprehensive benchmarking of cooperation strategies.
Specifically, we conduct comparison experiments on two challenging maps, \textit{10m vs 11m} (\textit{Hard}, \textit{homogeneous} and \textit{asymmetric}) and \textit{MMM2} (\textit{Super Hard}, \textit{heterogeneous} and \textit{asymmetric}).\looseness=-1

\noindent \textbf{Google Research Football}. GRF \cite{kurach2020google} is an open-source research environment designed for MARL algorithm evaluation in a simulated football setting. 
In GRF, agents play different roles within a football team (e.g., forwards, wingers, etc.), demonstrating evident role heterogeneity. 
We utilize the \textit{academy pass and shoot with keeper}, \textit{academy counterattack easy}, and \textit{academy 3 vs 1 with keeper} scenarios for algorithm evaluation.\looseness=-1

\noindent \textbf{Multi-Agent MuJoCo}. 
MA MuJoCo \cite{de2020deep} contains a variety of multi-agent continuous control tasks where individual agents control the joints of biomimetic robot entities and coordinate to facilitate specific behavior.
Built on the MuJoCo physics engine, MA MuJoCo's modular architecture allows for easy customization of the environments and agents' behavior, facilitating the simulation of complex interactions among agents trained by MARL algorithms.  
We evaluate the proposed method using the \textit{Ant-v2} scenario with two different configurations: \textit{8×1 agent Ant} and \textit{4×2 agent Ant}.\looseness=-1

\begin{figure*}[h]
  \centering
  \begin{subfigure}[b]{0.33\textwidth}
  \centering
    \includegraphics[width=0.85\textwidth]{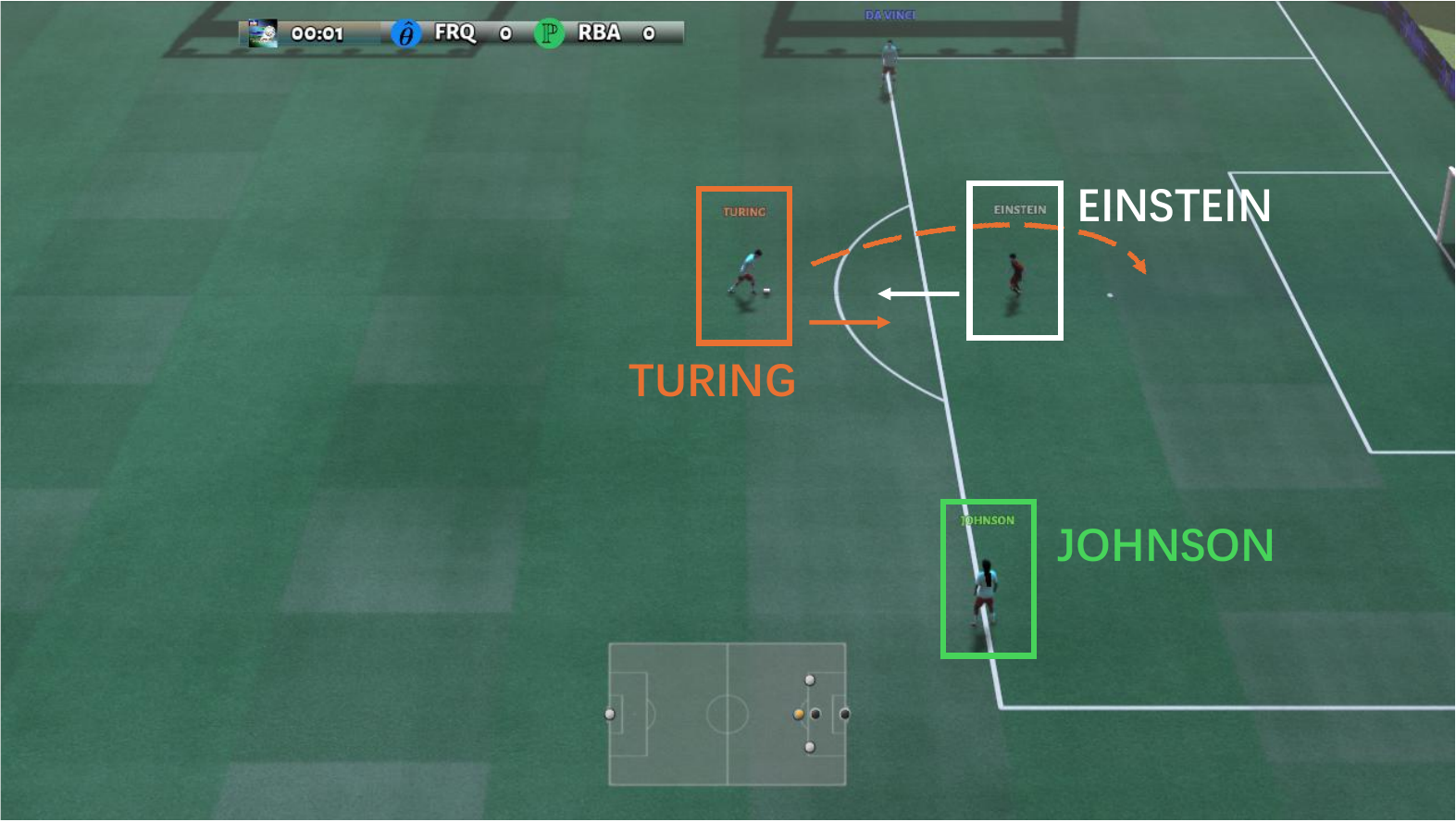}
    \caption{TURING attempts to bypass EINSTEIN.}
    \label{fig:MAT-1}
  \end{subfigure}
  \hfill
  \begin{subfigure}[b]{0.33\textwidth}
  \centering
    \includegraphics[width=0.85\textwidth]{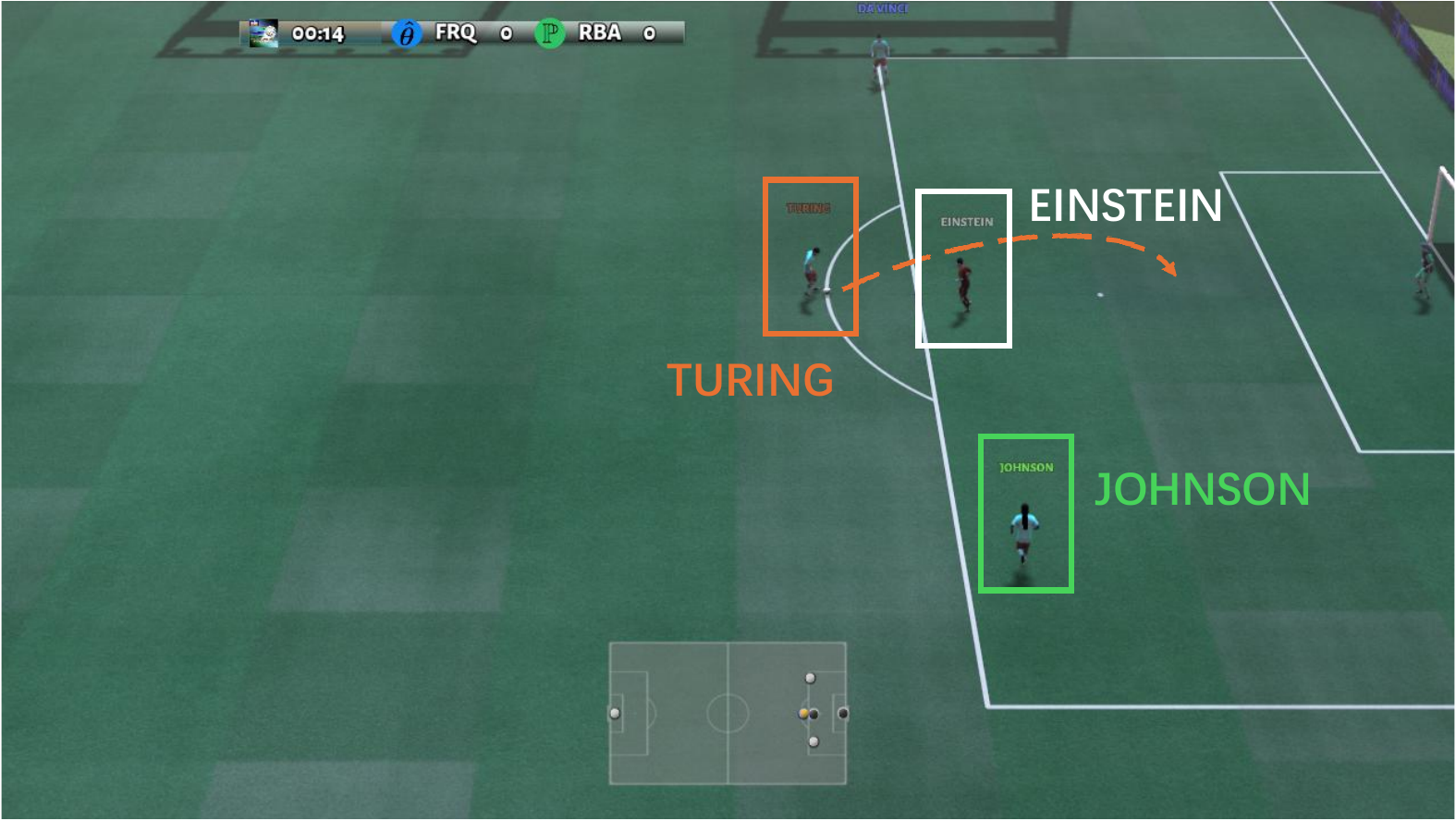}
    \caption{EINSTEIN mounts a frontal defense.}
    \label{fig:MAT-2}
  \end{subfigure}
  \hfill
  \begin{subfigure}[b]{0.33\textwidth}
  \centering
    \includegraphics[width=0.85\textwidth]{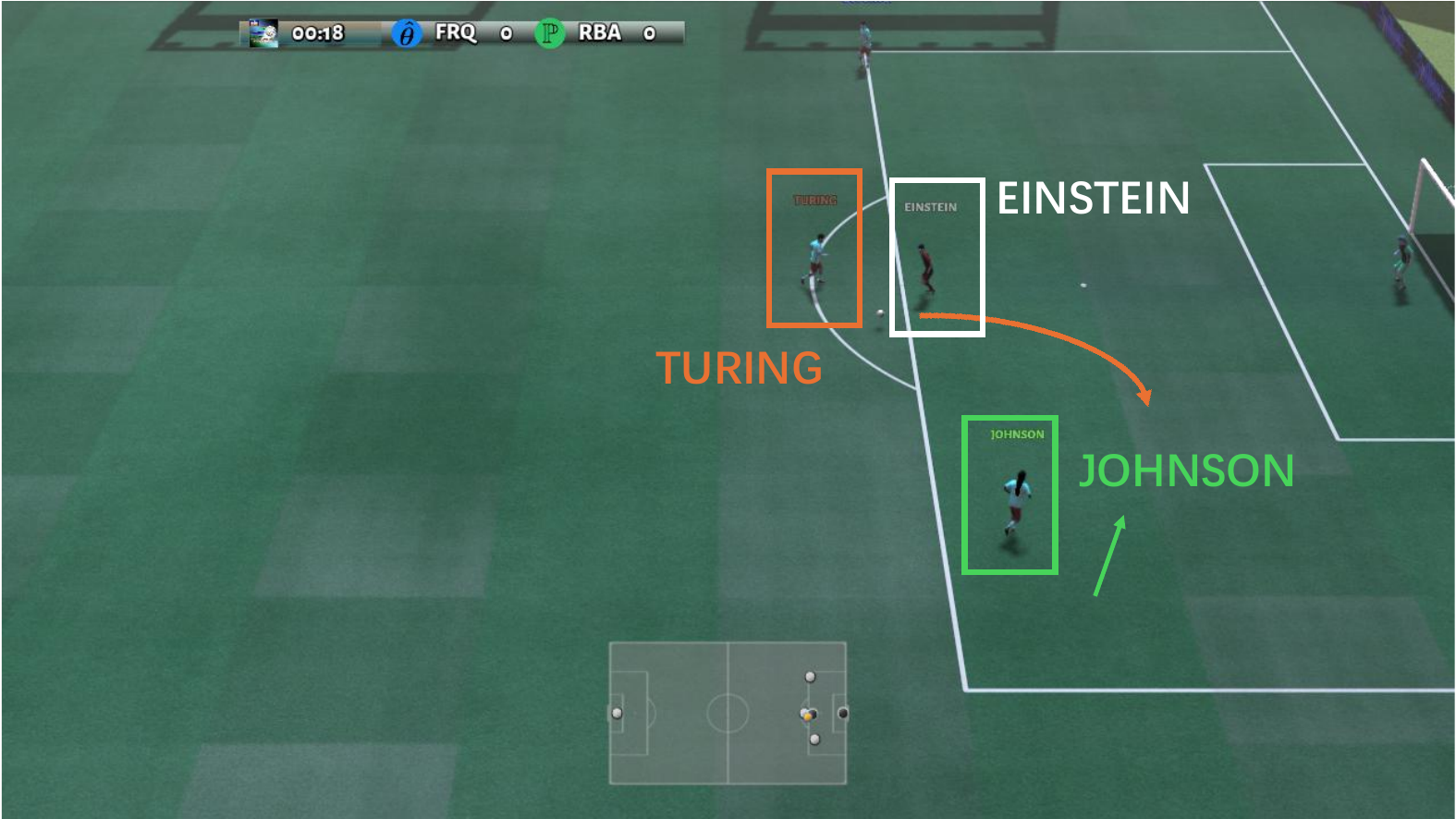}
    \caption{TURING opts for a pass to JOHNSON.}
    \label{fig:MAT-3}
  \end{subfigure}
  \\
  \begin{subfigure}[b]{0.33\textwidth}
  \centering
    \includegraphics[width=0.85\textwidth]{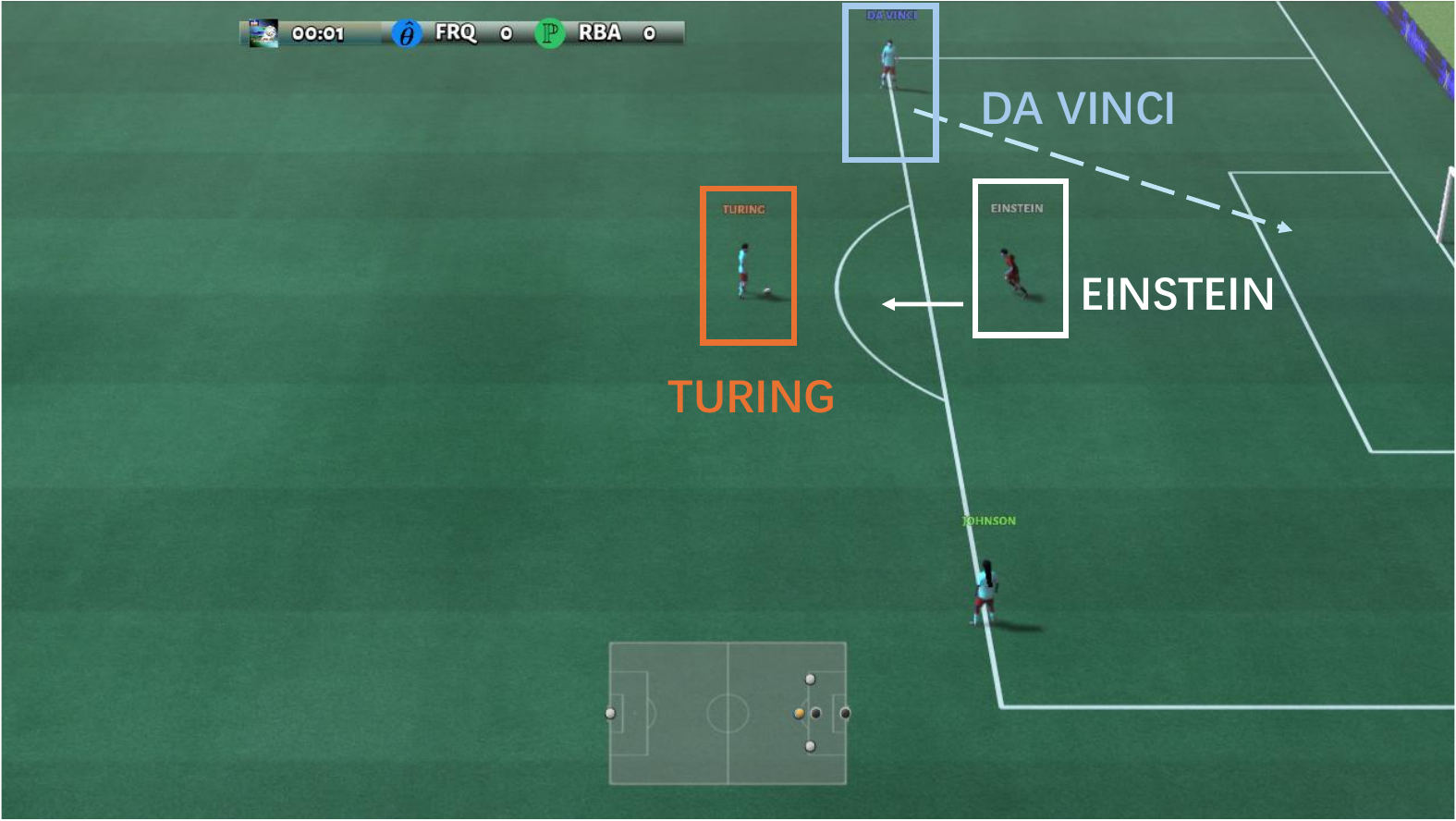}
    \caption{TURING maintains the ball waiting.}
    \label{fig:PMAT-1}
  \end{subfigure}
  \hfill
  \begin{subfigure}[b]{0.33\textwidth}
  \centering
    \includegraphics[width=0.85\textwidth]{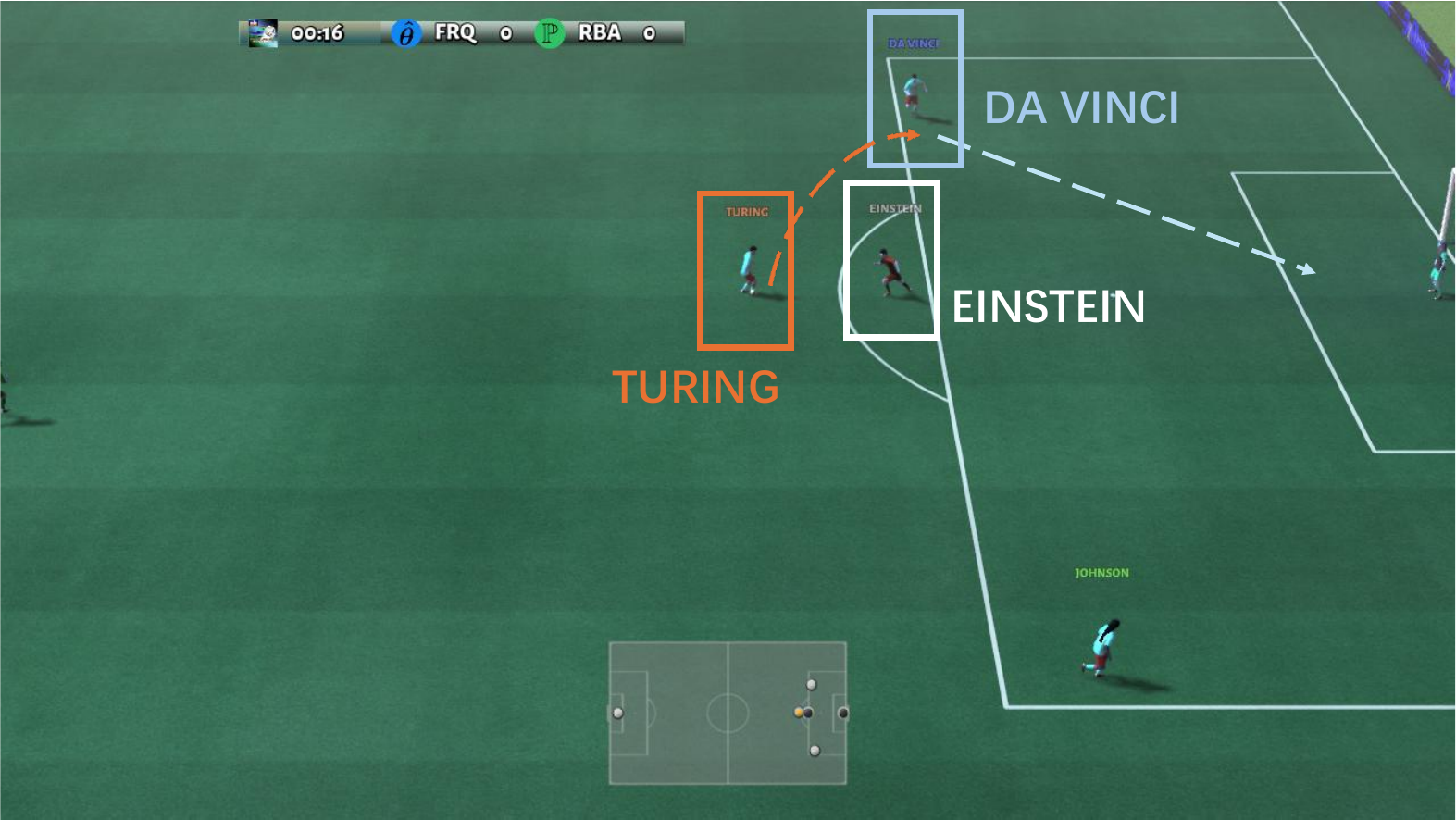}
    \caption{TURING passes the ball to DA VINCI.}
    \label{fig:PMAT-2}
  \end{subfigure}
  \hfill
  \begin{subfigure}[b]{0.33\textwidth}
  \centering
    \includegraphics[width=0.85\textwidth]{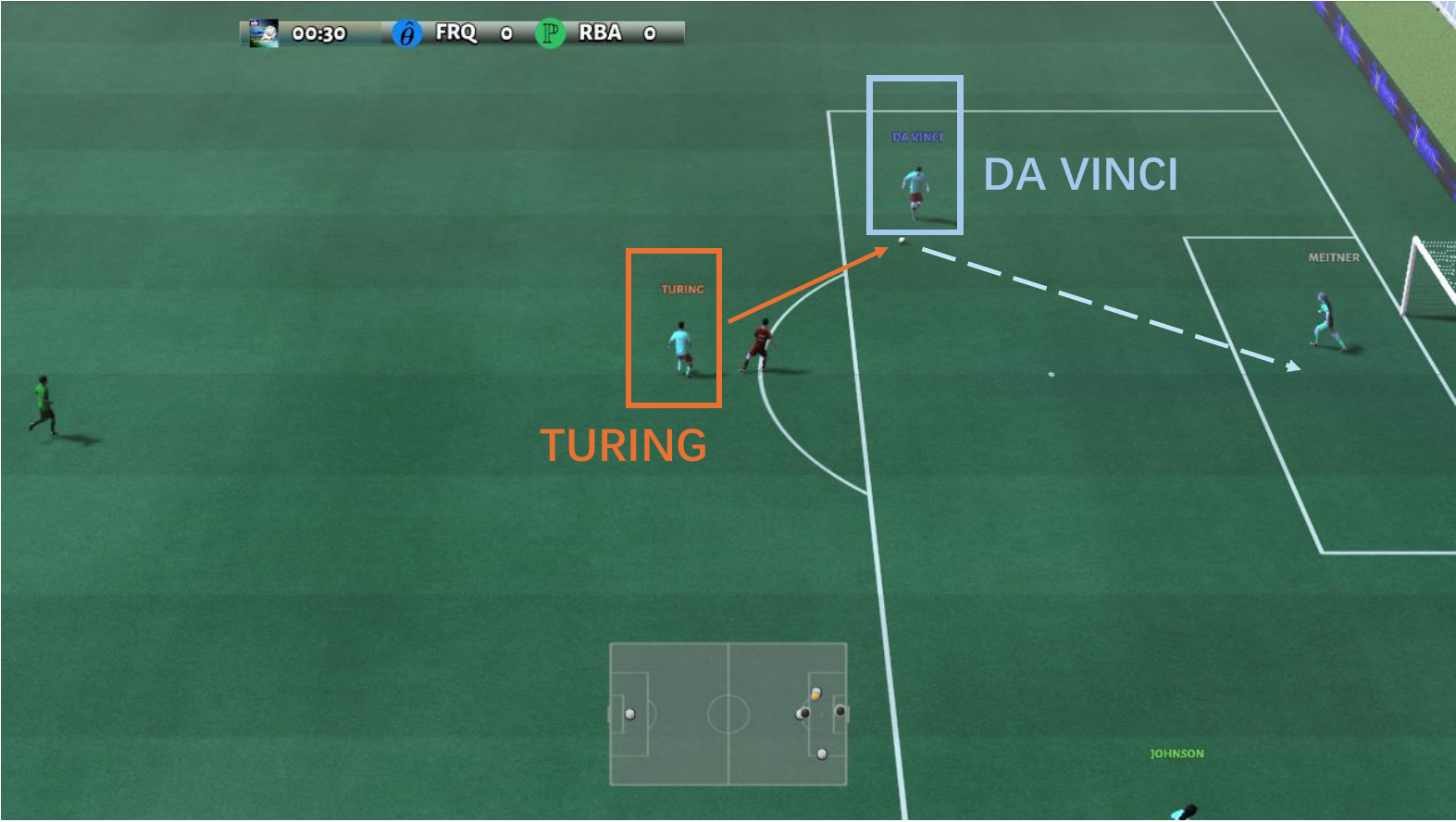}
    \caption{DA VINCI gets the ball and shoots.}
    \label{fig:PMAT-3}
  \end{subfigure}
  \caption{Different cooperation strategies of agents trained by MAT and PMAT in a Google Research Football setting. \Cref{fig:MAT-1,fig:MAT-2,fig:MAT-3} illustrate the behavior of agents trained by MAT. \Cref{fig:PMAT-1,fig:PMAT-2,fig:PMAT-3} illustrate the behavior of agents trained by PMAT. The solid lines represent the current action, whereas the dashed lines denote the predicted intention of the following action.\looseness=-1}
  \label{fig:pmat_mat_grf}
  \vspace{-8pt}
\end{figure*}

\subsection{Experimental Results}
\textbf{StarCraft II Multi-Agent Challenge.} The comparison results of all methods in the SMAC domain are shown in \Cref{fig:10m_vs_11m} and \Cref{fig:MMM2}.
Specifically, for the \textit{10m vs 11m} map, PMAT achieves a winrate of 99.4\%, outperforming the baseline methods MAT (97.5\%), MAPPO (75.0\%), and HAPPO (93.1\%). For the \textit{MMM2} map, PMAT achieves a winrate of 85.0\%, consistently surpassing the baseline methods MAT (75.0\%), MAPPO (58.8\%), and HAPPO (61.9\%). 
The experimental results indicate that the performance enhancement of PMAT over MAT tends to be more pronounced in heterogeneous scenarios (e.g., \textit{MMM2}) than in homogeneous scenarios (e.g., \textit{10m vs 11m}). 
This observation broadly corroborates our hypothesis in \Cref{sec:intro}. Specifically, as the heterogeneity intensifies the sequential dependencies among agents within such tasks, effective management of these dependencies significantly improves the coordination efficiency.
In addition, given that both of the selected maps contain ten ally agents, the experimental results in the SMAC domain can also validate the scalability of the proposed method.\looseness=-1

\noindent \textbf{Google Research Football.} The comparison results of all methods in the GRF domain are illustrated in \Cref{fig:academy_pass_and_shoot_with_keeper} and \Cref{fig:academy_counterattack_easy}. 
PMAT demonstrates 0.964 in average episode scores in the \textit{academy pass and shoot with keeper} scenario, which outperforms MAT (0.947), MAPPO (0.948), and HAPPO (0.890). In the \textit{academy counterattack easy} scenario, PMAT demonstrates 0.899 in average episode scores, consistently surpassing MAT (0.780), MAPPO (0.588), and HAPPO (0.533). 
As illustrated in \Cref{fig:experimental results}, while introducing the scoring network elevates the training cost and marginally degrades its performance in the early stages, PMAT ultimately surpasses MAT with the advancement of training. 
The experimental results in the GRF domain demonstrate that PMAT achieves superior performance in cooperation tasks that exhibit evident role heterogeneity.
Moreover, the performance curve of MAT exhibits more pronounced fluctuations than PMAT in some scenarios (e.g., \textit{academy counterattack easy}),  further validating the efficacy of the proposed AGPS mechanism in enhancing the stability of MARL algorithms.\looseness=-1

\noindent \textbf{Multi-Agent MuJoCo.} 
The comparison results of all methods in the MA MuJoCo domain are presented in \Cref{fig:8×1 agent Ant} and \Cref{fig:4×2 agent Ant}. 
It can be observed that, equipped with AGPS, PMAT consistently outperforms the baselines in both the \textit{8×1 agent Ant} and the \textit{4×2 agent Ant} scenarios. 
As illustrated in \Cref{fig:experimental results}, while PMAT initially exhibits comparable task performance to MAT, it progressively outperforms MAT as training proceeds, which demonstrates the effectiveness of the proposed method.
Additionally, it can be observed that in both of the \textit{Ant-v2} scenarios, the sequential decision-making MARL algorithms like MAT and PMAT exhibit distinct advantages in task performance over the simultaneous decision-making MARL algorithms like MAPPO and HAPPO.
We attribute this phenomenon to the lack of essential information regarding previous agents' decisions, which exacerbates coordination challenges as the torques applied by different agents may counteract each other, thus leading to degradation in overall task performance.\looseness=-1

\vspace{-5pt}
\subsection{Ablation Study}
In this section, we present an ablation study to further evaluate the advantage-based scoring mechanism adopted in AGPS across the SMAC, GRF, and MA MuJoCo benchmarks. Specifically, in addition to MAT, we introduce another baseline, namely the randomized MAT (abbreviated as rMAT in this section), which adopts a fully randomized ordering strategy to determine the action generation order based on MAT. We summarize the experimental results in \Cref{fig:ablation results}, where PMAT exhibits superior performance compared with both MAT and rMAT in all displayed tasks. It can be observed that the randomized ordering strategy introduces instability since rMAT exhibits pronounced variance in some scenarios (e.g., \textit{academy counterattack easy}), resulting in performance degradation. In contrast, the integration of AGPS effectively enhances both stability and monotonicity in task performance improvement, validating the effectiveness of the advantage-based scoring mechanism.\looseness=-1

\vspace{-5pt}
\subsection{Case Study}

In this section, we aim to analyze the distinct coordination strategies of agents trained by MAT and PMAT. Specifically, we conduct a fine-grained case study on agents' coordination behavior in the \textit{academy 3 vs 1 with keeper} scenario of GRF, as illustrated in \Cref{fig:pmat_mat_grf}. It can be observed that given identical initial conditions (our player TURING holds the ball, directly facing the defense of the opponent player EINSTEIN), agents trained by MAT adopt a strategy wherein TURING attempts to dribble past EINSTEIN and shoot (\Cref{fig:MAT-1,fig:MAT-2,fig:MAT-3}). In contrast, agents trained by PMAT adopt a different strategy (\Cref{fig:PMAT-1,fig:PMAT-2,fig:PMAT-3}), wherein TURING directly awaits an opportune moment to pass the ball to his teammate DA VINCI, who enjoys a superior shooting angle unimpeded by direct opposition. Compared with the former strategy wherein agents make decisions more based on their own observations, the latter strategy evaluates the significance of local observations within a multi-agent team and allows agents who hold advantageous observations to make decisions first (DA VINCI, shoot), followed by proactive coordination of other agents (TURING, pass the ball to DA VINCI). 
Generally, the comparison between the aforementioned coordination strategies in this section offers an intuitive validation for the necessity of agent decision order optimization in MARL, demonstrating its effectiveness in promoting cooperative behavior among agents.\looseness=-1

\section{CONCLUSION}
In this work, we propose \textbf{Action Generation with Plackett-Luce Sampling (AGPS)}, a sequential decision-making mechanism in MARL. AGPS assigns decision credits to individual agents within a multi-agent team and significantly facilitates joint policy improvement by providing finer-grained supervision for decision order optimization. Integrating AGPS with the Multi-Agent Transformer, we propose the \textbf{Prioritized Multi-agent Transformer (PMAT)}, a sequential decision-making MARL algorithm with optimized decision-ordering. Extensive experiments across various benchmarks showcase the effectiveness of AGPS as well as the superiority of PMAT in learning efficiency and task performance over several strong baselines. 
For future work, we plan to further investigate the effectiveness of AGPS by integrating it with a broader context of MARL algorithms. Besides, we will also evaluate the applicability of P-L sampling in larger-scale multi-agent systems.\looseness=-1



\clearpage

\begin{acks}
This work is supported by the National Natural Science Foundation of China (Grant Nos. 62106278, 91948303-1, 611803375, 12002380, 62101575) and the National Key Research and Development Program of China (Grant No. 2021ZD0140301).
Team from Shanghai Jiao Tong University is supported by the National Key R\&D Program of China (2022ZD0114804).
Muning Wen and Xihuai Wang are both supported by the Wen-Tsun Wu AI Honorary Doctoral Scholarship from AI Institute, Shanghai Jiao Tong University. 
\end{acks}



\bibliographystyle{ACM-Reference-Format}
\balance
\bibliography{sample}

\clearpage

\appendix

\section{Experimental Details}
\subsection{Introduction for Environments}
\begin{figure}[h]
  \centering
  \begin{subfigure}[b]{0.40\textwidth}
    \includegraphics[width=\textwidth]{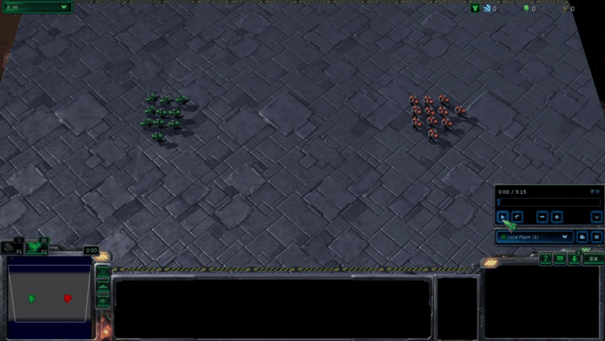}
    \caption{StarCraft II Multi-Agent Challenge}
    \label{fig:smac_screenshot}
  \end{subfigure}
  \hfill
  \begin{subfigure}[b]{0.40\textwidth}
    \includegraphics[width=\textwidth]{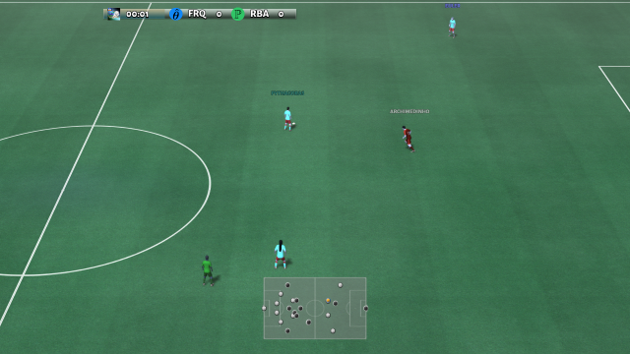}
    \caption{Google Research Football}
    \label{fig:grf_screenshot}
  \end{subfigure}
  \hfill
  \begin{subfigure}[b]{0.40\textwidth}
    \includegraphics[width=\textwidth]{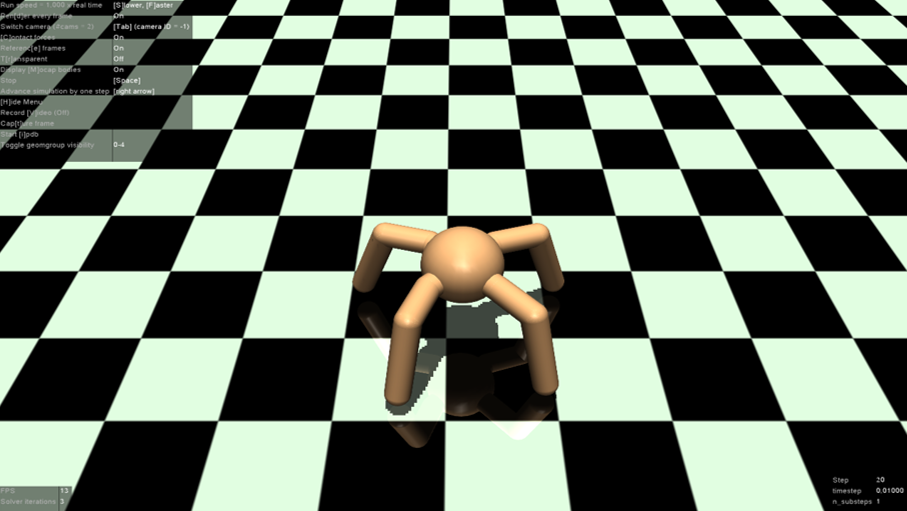}
    \caption{Multi-Agent MuJoCo}
    \label{fig:mujoco_screenshot}
  \end{subfigure}
  \vspace{-8pt}
  \caption{Screenshots of the SMAC, GRF, and MA MuJoCo environments used in this paper for reference.}
  \label{fig:screenshots}
  \vspace{-8pt}
\end{figure}

This paper utilizes three widely-used MARL benchmarks: StarCraft II Multi-Agent Challenge (SMAC) \cite{samvelyan2019starcraft}, Google Research Football (GRF) \cite{kurach2020google}, and Multi-Agent MuJoCo (MA MuJoCo) \cite{de2020deep}. Screenshots of these environments are provided in \Cref{fig:screenshots} for reference.

\subsection{Scenario Description}
\label{sec:description}
This section presents detailed descriptions of the scenarios adopted in experiments. Note that the descriptions are taken from the official documentation of the corresponding benchmarks.\looseness=-1

\noindent \textbf{Descriptions of the SMAC scenarios adopted in experiments}. 
\textit{10m vs 11m}: 10 Marines of ally units versus 11 Marines of enemy units. The difficulty level is \textit{hard}. \textit{MMM2}: 1 Medivac, 2 Marauders, and 7 Marines of ally units versus 1 Medivac, 3 Marauders, and 8 Marines of enemy units. The difficulty level is \textit{super hard}.\looseness=-1

\noindent \textbf{Descriptions of the GRF scenarios adopted in experiments}.
\textit{academy pass and shoot with keeper}: Two of our players try to score from the edge of the box (one is on the side with the ball next to a defender, the other is at the center, unmarked and facing the opponent keeper). \textit{academy counterattack easy}: Four of our players play against one opponent player with all the remaining players of both teams running back toward the ball. \textit{academy 3 vs 1 with keeper}: Three of our players try to score from the edge of the box, one on each side and the other at the center. The player at the center initially has the ball and faces the opponent defender.\looseness=-1

\noindent \textbf{Descriptions of the MA MuJoCo scenario adopted in experiments}. 
\textit{Ant-v2}: Based on the environment introduced by \cite{schulman2015high}, this environment simulates a 3D ant-like robot entity consisting of one torso (rotational) with four attached legs, each leg having two joints. The optimization goal is to coordinate the four legs by applying torques on the eight hinges connecting the two body parts of each leg and the torso so that the robot entity can move in the correct direction. There are nine body parts and eight hinges in total.\looseness=-1

\subsection{Implementations}
\label{sec:implementation}

In this study, all experiments were conducted using an NVIDIA 3090 GPU. The programming environment was established with Python 3.9 and PyTorch 2.3.1. To ensure robustness, each experiment was repeated with three distinct random seeds, and the results were visualized using the mean values of the performance metrics (median value for the SMAC environment).
Additionally, it should be noted that minor adjustments were made to the implementations of the SMAC environment due to identified technical inconsistencies\footnote{For further details, please refer to \url{https://github.com/oxwhirl/smac/issues/72}.} in the reward function of the original SMAC environment.\looseness=-1

\subsection{Hyperparameter Settings}
\label{sec:hyperparameters}
This section presents the detailed hyperparameter settings for different algorithms adopted in experiments.

Specifically, \Cref{tab:chpSMAC} shows the common hyperparameters we used in SMAC experiments.
\Cref{tab:chpgrf} shows the common hyperparameters we used in GRF experiments.
\Cref{tab:chpMAM} shows the common hyperparameters we used in MA MuJoCo experiments.\looseness=-1

\Cref{tab:differentSMAC} shows the different hyperparameters we used for MAPPO and HAPPO in SMAC experiments.
\Cref{tab:differentSMAC_MAT} shows the different hyperparameters we used for MAT and PMAT in SMAC experiments.
\Cref{tab:differentGRF} shows the different hyperparameters we used for all methods in GRF experiments.
\Cref{tab:differentGRFpmat} shows the customized hyperparameters we used for PMAT in GRF experiments.
\Cref{tab:differentMAM} shows the different hyperparameters we used for all methods in MA MuJoCo experiments.
\Cref{tab:differentMAMpmat} shows the customized hyperparameters we used for PMAT in MA MuJoCo experiments.\looseness=-1

\begin{table*}[h]
\centering
\caption{Common hyperparameters used for PMAT, MAT, MAPPO and HAPPO in the SMAC domain.}\label{tab:chpSMAC}
\begin{tabular}{cc|cc}
    \toprule
    Hyperparameters & Value & Hyperparameters & Value \\
    \midrule
    critic lr & 5e-4 & actor lr & 5e-4 \\
    gain & 0.01 & optim eps & 1e-5 \\
    training threads & 16 & num mini-batch & 1 \\
    entropy coef & 0.01 & max grad norm & 10 \\
    optimizer & Adam & hidden layer dim & 64 \\
    use gae & True & episode length & 100 \\
    batch size & 3200 & use huber loss & True \\
    rollout threads & 32 & environment steps & 5e6 \\
    \bottomrule
\end{tabular}
\end{table*}

\begin{table*}[h]
\centering
\caption{Common hyperparameters used for all methods in the Google Research Football domain.}\label{tab:chpgrf}
\begin{tabular}{cc|cc}
    \toprule
    Hyperparameters & value & Hyperparameters & value \\
    \midrule
    gamma & 0.99 & stacked frames & 1 \\
    gain & 0.01 & optim eps & 1e-5 \\
    training threads & 16 & num mini-batch & 1 \\
    entropy coef & 0.01 & max grad norm & 0.5 \\
    optimizer & Adam & hidden layer dim & 64 \\
    rollout threads & 20 & episode length & 200 \\
    batch size & 4000 & use huber loss & True \\
    \bottomrule
\end{tabular}
\end{table*}

\begin{table*}[h]
\centering
\caption{Common hyperparameters used for all methods in the Multi-Agent MuJoCo domain.}\label{tab:chpMAM}
\begin{tabular}{cc|cc}
    \toprule
    Hyperparameters & Value & Hyperparameters & Value \\
    \midrule
    gamma & 0.99 & stacked frames & 1 \\
    gain & 0.01 & optim eps & 1e-5 \\
    training threads & 16 & num mini-batch & 40 \\
    entropy coef & 0.001 & max grad norm & 0.5 \\
    optimizer & Adam & hidden layer dim & 64 \\
    rollout threads & 40 & episode length & 100 \\
    batch size & 4000 & use huber loss & True \\
    \bottomrule
\end{tabular}
\end{table*}

\begin{table*}[h]
\centering
\caption{Different hyperparameters used for MAPPO and HAPPO in the SMAC domain.}\label{tab:differentSMAC}
\begin{tabular}{c|ccc}
    \toprule
    maps & ppo epochs & ppo clip & num hidden layers \\
    \midrule
    \textit{10m vs 11m} & 10 & 0.2 & 2 \\
    \textit{MMM2} & 5 & 0.2 & 1 \\
    \midrule
    maps & stacked frames & network & $\gamma$ \\
    \midrule
    \textit{10m vs 11m} & 1 & mlp & 0.95 \\
    \textit{MMM2} & 1 & mlp & 0.95 \\
    \bottomrule
\end{tabular}
\end{table*}

\begin{table*}[h]
\centering
\caption{Different hyperparameters used for PMAT and MAT in the SMAC domain.}\label{tab:differentSMAC_MAT}
\begin{tabular}{c|ccc}
    \toprule
    maps & ppo epochs & ppo clip & num blocks \\
    \hline
    \textit{10m vs 11m} & 10 & 0.05 & 1 \\
    \textit{MMM2} & 5 & 0.05 & 1 \\
    \midrule
    maps & num heads & stacked frames & $\gamma$ \\
    \hline
    \textit{10m vs 11m} & 1 & 1 & 0.99 \\
    \textit{MMM2} & 1 & 1 & 0.99 \\
    \midrule
    maps & ranking loss coef & num ranking layers &  \\
    \hline
    \textit{10m vs 11m} & 1e-4 | / & 2 | / &  \\
    \textit{MMM2} & 1e-1 | / & 2 | / &  \\
    \bottomrule
\end{tabular}
\end{table*}

\begin{table*}[h]
\centering
\caption{Different hyperparameters used for all methods in the Google Research Football domain.}\label{tab:differentGRF}
\begin{tabular}{c|cccc}
    \toprule
    Hyperparameters & PMAT & MAT & MAPPO & HAPPO \\
    \midrule
    ppo epochs & 10 & 10 & 5 & 5 \\
    ppo clip & 0.05 & 0.05 & 0.2 & 0.2 \\
    num hidden layers & / & / & 2 & 2 \\
    num blocks & 1 & 1 & / & / \\
    num heads & 1 & 1 & / & / \\
    \bottomrule
\end{tabular}
\end{table*}

\begin{table*}[h]
\centering
\caption{Customized hyperparameters used for PMAT in the Google Research Football domain.}\label{tab:differentGRFpmat}
\begin{tabular}{c|cc}
    \toprule
    Scenarios & num ranking layers & ranking loss coefficient \\
    \midrule
    \textit{academy pass and shoot with keeper} & 2 & 1e-4 \\
    \textit{academy counterattack easy} & 3 & 1e-1 \\
    \textit{academy 3 vs 1 with keeper} & 3 & 1e-1 \\
    \bottomrule
\end{tabular}
\end{table*}

\begin{table*}[h]
\centering
\caption{Different hyperparameters used for all methods in the Multi-Agent MuJoCo domain.}\label{tab:differentMAM}
\begin{tabular}{c|cccc}
    \toprule
    Hyperparameters & PMAT & MAT & MAPPO & HAPPO \\
    \midrule
    critic lr & 5e-5 & 5e-5 & 5e-3 & 5e-3 \\
    actor lr & 5e-5 & 5e-5 & 5e-6 & 5e-6 \\
    ppo epochs & 10 & 10 & 5 & 5 \\
    ppo clip & 0.05 & 0.05 & 0.2 & 0.2 \\
    num hidden layers & / & / & 2 & 2 \\
    num blocks & 1 & 1 & / & / \\
    num heads & 1 & 1 & / & / \\
    \bottomrule
\end{tabular}
\end{table*}

\begin{table*}[h]
\centering
\caption{Customized hyperparameters used for PMAT in the Multi-Agent MuJoCo domain.}\label{tab:differentMAMpmat}
\begin{tabular}{c|cc}
    \toprule
    Scenarios & num ranking layers & ranking loss coefficient \\
    \midrule
    \textit{8×1 agent Ant-v2} & 3 & 1e-4 \\
    \textit{4×2 agent Ant-v2} & 3 & 1e-3 \\
    \bottomrule
\end{tabular}
\end{table*}


\end{document}